\documentclass{article} 
\usepackage{iclr2025_conference,times}

\usepackage{amsmath,amsfonts,bm}

\def\eqref#1{equation~\ref{#1}}

\def\1{\bm{1}}

\def\vx{{\bm{x}}}

\DeclareMathAlphabet{\mathsfit}{\encodingdefault}{\sfdefault}{m}{sl}
\SetMathAlphabet{\mathsfit}{bold}{\encodingdefault}{\sfdefault}{bx}{n}

\usepackage{hyperref}
\usepackage[capitalise]{cleveref}
\usepackage{url}
\usepackage{multirow}
\usepackage{graphicx}
\usepackage{subfigure}
\usepackage{adjustbox}
\usepackage{amsmath,amsfonts,bm}
\usepackage{pifont}
\usepackage{xcolor}

\newtheorem{theorem}{Theorem}
\iclrfinalcopy
\title{Sinc Kolmogorov-Arnold network and its application for solving PDEs with singularities}

\makeatletter
\let\@fnsymbol\@arabic
\makeatother
\author{Tianchi Yu \thanks{Skolkovo Institute of Science and Technology, corresponding author: \textit{tianchi.yu@skoltech.ru}} \\
\And
Jingwei Qiu \thanks{Southern University of Science and Technology}\\ 
\And
Jiang Yang \thanks{SUSTech International Center for Mathematics \& National Center for Applied Mathematics Shenzhen (NCAMS)}
\And
Ivan Oseledets \thanks{Skolkovo Institute of Science and Technology; AIRI} \\
}

\begin{document}

\maketitle

\begin{abstract}
In this paper, we propose to use Sinc interpolation in the context of Kolmogorov-Arnold Networks, neural networks with learnable activation functions, which recently gained attention as alternatives to Multilayer Perceptron. Many different function representations have already been tried, but we show that Sinc interpolation proposes a viable alternative, since it is known in numerical analysis to effectively represent both smooth functions and functions with singularities. This is important not only for function approximation but also for solving the partial differential equations with physics-informed neural networks. Through a series of experiments, we show that SincKANs provide better results in almost all of the examples we have considered.
\end{abstract}
\section{Introduction}\label{section: introduction}
Multilayer perceptron (MLP) is a classical neural network consisting of fully connected layers with a chosen nonlinear activation function, which is a superposition of simple functions. The classical Kolmogorov-Arnold representation theorem \cite{kolmogorov1961representation,arnol1959representation} states that every function can be represented as a superposition of function of at most $2$ variables, motivating the research for learnable activation functions.

Kolmogorov's Spline Network (KSN) \cite{igelnik2003kolmogorov} is a two-layer framework using splines as the learnable activation functions. Recently, Kolmogorov-Arnold Networks (KANs) \cite{liu2024kan} sparkled a new wave of attention to those approaches, by proposing a multilayer variant of KSN. Basically, any successful basis to represent univariate functions can provide a new variant of KAN. Many well-known methods have already been investigated including wavelet \cite{bozorgasl2024wav,seydi2024unveiling}, Fourier series \cite{xu2024fourierkan}, finite basis \cite{howard2024finite}, Jacobi basis functions \cite{aghaei2024fkan}, polynomial basis functions \cite{seydi2024exploring}, rational functions \cite{aghaei2024rkan} and Chebyshev polynomials \cite{ss2024chebyshev,shukla2024comprehensive}.  

We propose to use Sinc interpolation (the Sinc function is defined in \cref{eq:sinc}) which is a very efficient and well-studied method for function interpolation, especially 1D problems \cite{stenger2016handbook}. To our knowledge, it has not been studied in the context of KANs. We argue that the the cubic spline interpolation used in KANs should be replaced by the Sinc interpolation, because splines are particularly good for the approximation of analytic functions without singularities which MLP is also good at, while Sinc methods excel for problems with singularities, for boundary-layer problems, and for problems over infinite or semi-infinite range \cite{stenger2012numerical}. Herein, utilizing Sinc functions can improve the accuracy and generalization of KANs, and make KANs distinguishing and competitive, especially in solving aforementioned mathematical problems in machine learning. We will confirm our hypothesis by numerical experiments.

Physics-informed neural networks (PINNs) \cite{lagaris1998artificial,raissi2019physics} are a method used to solve partial differential equations (PDEs) by integrating physical laws with neural networks in machine learning. The use of Kolmogorov-Arnold Networks (KANs) in PINNs has been explored and is referred to as Physics-Informed Kolmogorov-Arnold Networks (PIKANs) \cite{rigas2024adaptive,wang2024kolmogorov}. Due to the high similarity between KAN and MLP, PIKANs inherit several advantages of PINNs, such as overcoming the curse of dimensionality (CoD) \cite{wojtowytsch2020can,han2018solving}, handling imperfect data \cite{karniadakis2021physics}, and performing interpolation \cite{sliwinski2023mean}. PINNs have diverse applications, including fluid dynamics \cite{raissi2020hidden,jin2021nsfnets,kashefi2022physics}, quantum mechanical systems \cite{jin2022physics}, surface physics \cite{fang2019physics}, electric power systems \cite{nellikkath2022physics}, and biological systems \cite{yazdani2020systems}. However, they also face challenges such as spectral bias \cite{xu2019frequency,wang2022and}, error estimation \cite{fanaskov2024astral}, and scalability issues \cite{yao2023multiadam}.

In this paper, we introduce a novel network architecture called Sinc Kolmogorov-Arnold Networks (SincKANs). This approach leverages Sinc interpolation, which is particularly adept at approximating functions with singularities, to replace cubic interpolation in the learnable activation functions of KANs. The ability to handle singularities enables SincKAN to mitigate the spectral bias observed in PIKANs, thereby making PIKANs more robust and capable of solving PDEs that traditional PINNs may struggle with. Additionally, we conducted a series of experiments to validate SincKAN's interpolation capabilities and assess their performance as a replacement for MLP and KANs in PINNs.
Our specific contributions can be summarized as follows:
\begin{enumerate}
    \item We propose the Sinc Kolmogorov-Arnold Networks, a novel network that excels in handling singularities.
    \item We propose several approaches based on classical techniques of Sinc methods that can enhance the robustness and performance of SincKAN.
    \item We conducted a series of experiments to demonstrate the performance of SincKAN in approximating a function and PIKANs.
\end{enumerate}

The paper is structured as follows: 
In \cref{Section: Methods}, we briefly introduce the PINNs, discuss Sinc numerical methods, and provide a detailed explanation of SincKAN. In  \cref{Section: experiments} we compare our SincKAN with several networks including  MLP, modified MLP \cite{wang2021understanding}, KAN, ChebyKAN in several diverse benchmarks including smooth functions, discontinuous functions, and boundary layer problems. In \cref{Section: conclusion}, we conclude the paper and discuss the remaining limitations and directions for future research.
\section{Methods}\label{Section: Methods}
\subsection{Physics-informed neural networks (PINNs)}
We briefly review the physics-informed neural networks (PINNs) \cite{raissi2019physics} in the context of
inferring the solutions of PDEs. Generally, we consider time-dependent PDEs for $\boldsymbol{u}$ taking the form
\begin{equation}
\begin{aligned}
    & \partial_{t}\boldsymbol{u}+\mathcal{N}[\boldsymbol{u}]=0, \quad t \in[0, T],\ \boldsymbol{x} \in \Omega, \\
    & \boldsymbol{u}(0, \boldsymbol{x})=\boldsymbol{g}(\boldsymbol{x}), \quad \boldsymbol{x} \in \Omega, \\
    & \mathcal{B}[\boldsymbol{u}]=0, \quad t \in[0, T],\ \boldsymbol{x} \in \partial \Omega,
\end{aligned}\label{PDE}
\end{equation}
where $\mathcal{N}$ is the differential operator, $\Omega$ is the domain of grid points, and $\mathcal{B}$ is the boundary operator. When considering time-independent PDEs, $\partial_{t}\boldsymbol{u}\equiv 0$. 

The ambition of PINNs is to approximate the unknown solution $\boldsymbol{u}$ to the PDE system \cref{PDE}, by optimizing a neural network $\boldsymbol{u}^{\theta}$, where $\theta$ denotes the trainable parameters of the neural network. The constructed loss function is:
\begin{equation}\label{PINN}
\mathcal{L}(\theta)=\mathcal{L}_{i c}(\theta)+\mathcal{L}_{b c}(\theta)+\mathcal{L}_r(\theta) ,
\end{equation}
where
\begin{equation}
\begin{aligned}
& \mathcal{L}_r(\theta)=\frac{1}{N_r} \sum_{i=1}^{N_r}\left|\partial_{t}\boldsymbol{u}^\theta\left(t_r^i, \boldsymbol{x}_r^i\right)+\mathcal{N}\left[\boldsymbol{u}^\theta\right]\left(t_r^i, \boldsymbol{x}_r^i\right)\right|^2,\\
& \mathcal{L}_{i c}(\theta)=\frac{1}{N_{i c}} \sum_{i=1}^{N_{i c}}\left|\boldsymbol{u}^\theta\left(0, \boldsymbol{x}_{i c}^i\right)-\boldsymbol{g}\left(\boldsymbol{x}_{i c}^i\right)\right|^2, \\
& \mathcal{L}_{b c}(\theta)=\frac{1}{N_{b c}} \sum_{i=1}^{N_{b c}}\left|\mathcal{B}\left[\boldsymbol{u}^\theta\right]\left(t_{b c}^i, \boldsymbol{x}_{b c}^i\right)\right|^2, \\
\end{aligned}
\end{equation}
corresponds to the three equations in \cref{PDE} individually; $\boldsymbol{x}_{i c}^i,\boldsymbol{x}_{b c}^i,\boldsymbol{x}_{r}^i$ are the sampled points from the initial constraint, boundary constraint, and residual constraint, respectively; $N_{i c},N_{b c},N_{r}$ are the total number of sampled points for each constraint, correspondingly. Note that in \cite{raissi2019physics}, $u^\theta\left(x\right)=\mathbf{MLP}\left(x\right)$.
\subsection{Sinc numerical methods}\label{section: sinc_numerical_methods}
The Sinc function is defined as\footnote{In engineering, they define Sinc function as $\mathrm{Sinc}(x)=\frac{\sin(\pi x)}{\pi x}$}
\begin{equation}\label{eq:sinc}
    \mathrm{Sinc}(x)=\frac{\sin(x)}{x},
\end{equation}
the Sinc series $S(j, h)(x)$ used in Sinc numerical methods is defined by:
\begin{equation}\label{candidate function}
S(j, h)(x)=\frac{\sin [(\pi / h)(x-j h)]}{(\pi / h)(x-j h)},    
\end{equation}
then the Sinc approximation for a function $f$ defined on the real line $\mathbb{R}$ is given by
\begin{equation}\label{eq: sinc_approx}
f(x) \approx \sum_{j=-N}^N f(j h) S(j, h)(x), \quad x \in \mathbb{R},
\end{equation}
where $h$ is the step size with the optimal value $\sqrt{\pi d/\beta N}$ provided in \cref{Theorem: Stenger}, and $2N+1$ is the degree of Sinc series.

Thanks to Sinc function's beautiful properties including the equivalence of semidiscrete Fourier
transform \cite{trefethen2000spectral}, its approximation as a nascent delta function, etc., Sinc numerical methods have become a technique for solving a wide range of linear and nonlinear problems arising from scientific and engineering applications including heat transfer \cite{lippke1991analytical}, fluid mechanics \cite{abdella2015solving}, and solid mechanics \cite{abdella2009application}. But Sinc series are the orthogonal basis defined on $\left(-\infty,\infty\right)$ which is impractical for numerical methods. To use Sinc numerical methods, one should choose a proper coordinate transformation based on the computing domain $\left(a,b\right)$ and an optimal step size based on the target function $f$. However, manually changing the network to meet every specific problem is impractical and wasteful. In the following of this section, we will introduce current techniques used in Sinc numerical methods. Then in \cref{Section: sinckan}, we will unfold Sinc numerical methods to meet machine learning.

\textit{Convergence theorem}

\begin{theorem}\label{Theorem: Stenger} \cite{sugihara2004recent}

Assume $\alpha, \beta, d >0$, that

(1) $f$ belongs to $H^1\left(\mathcal{D}_d\right)$, where $H^1$ is the Hardy space and $\mathcal{D}_d=\{z\in \mathbb{C} \ | \ \left|\Im z\right|<d\}$;

(2) $f$ decays exponentially on the real line, that is, $|f(x)| \leq \alpha \exp (-\beta|x|), \ \forall x \in \mathbb{R}$.

Then we have

\begin{equation}
\sup _{-\infty<x<\infty}\left|f(x)-\sum_{j=-N}^N f(j h) S(j, h)(x)\right| \leq C N^{1 / 2} \exp \left[-(\pi d \beta N)^{1 / 2}\right]    
\end{equation}
for some constant $C$, where the step size $h$ is taken as
\begin{equation}\label{optimal_h}
h=\left(\frac{\pi d}{\beta N}\right)^{1 / 2}.    
\end{equation}
\end{theorem}

\cref{Theorem: Stenger} indicates that the exponential convergence of Sinc approximation on the real line depends on the parameters $d,\beta,N$ which are determined by the target function $f$. Thus, in Sinc numerical methods, researchers set specific parameters for specific function $f$ \cite{sugihara2004recent,mohsen2017accurate} or set them by bisection \cite{richardson2011sinc}. Note that both approaches require the target function $f$, but in machine learning, $f$ is usually unknown.

\textit{On a general interval $(a,b)$}

Practical problems generally require approximating on an interval $(a,b)$ instead of the entire real line $\mathbb{R}$. To implement Sinc methods on general functions, we have to transform the interval $(a, b)$ to $\mathbb{R}$ with a properly selected coordinate transformation, \textit{i.e.} we define a transformation $x=\psi\left(\xi\right)$ such that $\psi: \left(-\infty,+\infty\right) \rightarrow \left(a,b\right)$. Then \cref{eq: sinc_approx} is replaced by 
\begin{equation}\label{eq: sinc_approx_trans}
f(\psi(\xi)) \approx \sum_{j=-N}^N f(\psi(j h)) S(j, h)(\xi), \quad-\infty<\xi<\infty,
\end{equation}
where $h$ is the step size with the optimal value $\sqrt{\pi d^\prime/\beta^\prime N}$ provided in \cref{Theorem: transformation}.
The following theorem states that \cref{Theorem: Stenger} still holds with some different $\alpha, \beta$, and $d$ after the coordinate transformation.
\begin{theorem} \label{Theorem: transformation} \cite{sugihara2004recent}

Assume that, for a variable transformation $x=\psi(\xi)$, the transformed function $f(\psi(\xi))$ satisfies assumptions 1 and 2 in \cref{Theorem: Stenger} with some $\alpha^\prime, \beta^\prime$ and $d^\prime$. Then we have

$$
\sup _{a \leq x \leq b}\left|f(x)-\sum_{j=-N}^N f(\psi(j h)) S(j, h)\left(\psi^{-1}(x)\right)\right| \leq C N^{1 / 2} \exp \left[-(\pi d^\prime \beta^\prime N)^{1 / 2}\right]
$$

for some $C$, where the step size $h$ is taken as

$$
h=\left(\frac{\pi d^\prime}{\beta^\prime N}\right)^{1 / 2}
$$
\end{theorem}
This theorem suggests the possibility that even a function $f$ with an end-point singularity can be approximated successfully by \cref{eq: sinc_approx_trans} with a suitable choice of transformation.

\begin{figure}[ht]
   \centering
    \subfigure[\label{sqrt_matlab}$f(x)=\sqrt{x}$]{\includegraphics[width=0.3\linewidth]{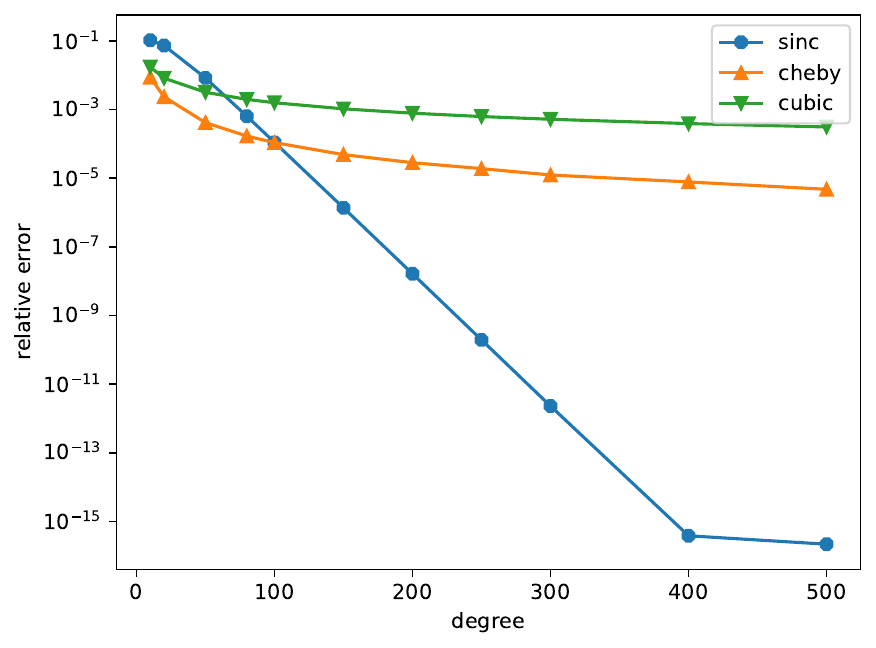}}
    \subfigure[\label{bl_matlab}$f(x)=e^{-10^5x}$]{\includegraphics[width=0.3\linewidth]{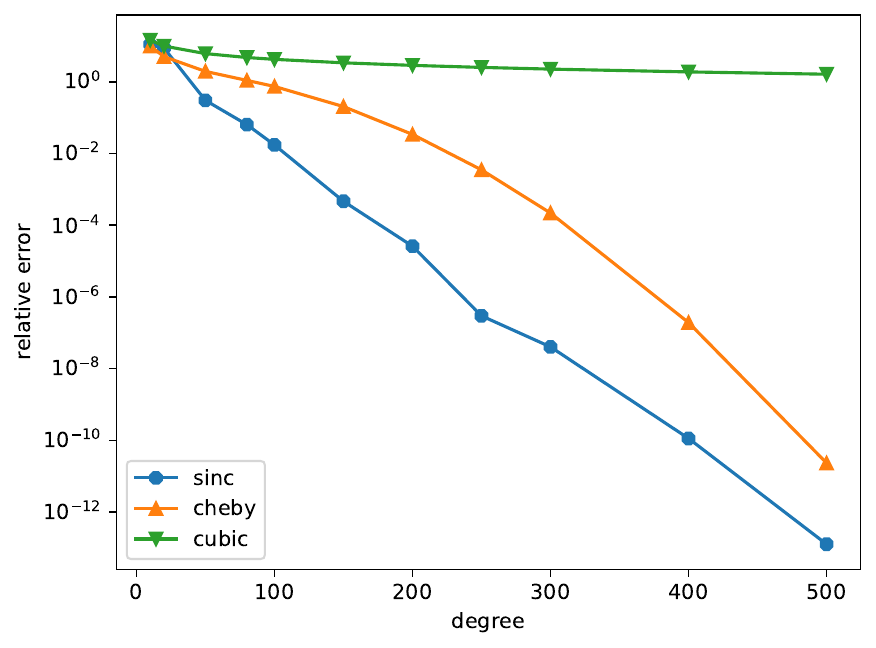}}
    \subfigure[\label{bl10000_solution}Comparison of \cref{bl_matlab}]{\includegraphics[width=0.3\linewidth]{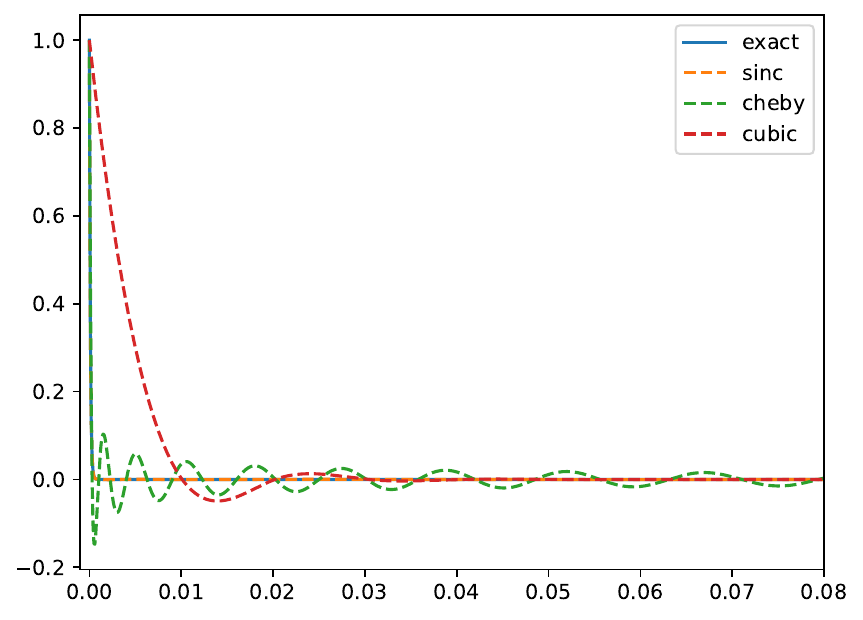}}
   \caption{\cref{sqrt_matlab} depicts the Sinc's merit of handling the end-point singularity while the Chebyshev and the spline converge slowly. \cref{bl_matlab} shows that, for the boundary layer functions that have high derivatives, Sinc converges exponentially while Chebyshev converges slowly at first. \cref{bl10000_solution} partially depicts the solution and the interpolations over the interval $[0,0.08]$, indicating that Sinc interpolation provides the most accurate approximation, while Chebyshev interpolation exhibits significant oscillations, and spline interpolation shows localized inaccuracies in certain regions}
    \label{fig:matlab}
\end{figure}

Furthermore, we empirically demonstrate the merits of Sinc methods in \cref{fig:matlab} via numerical results generated by Chebfun \cite{driscoll2014chebfun} for Chebyshev and cubic spline interpolation and Sincfun \cite{richardson2011sinc} for Sinc interpolation.

\subsection{Sinc Kolmogorov-Arnold Network (SincKAN)}\label{Section: sinckan}
Suppose $\mathbf{\Phi}=\{\phi_{p,q}\}$ is the matrix of univariate functions where $p= 1, 2, \dots n_{in}, q= 1, 2, \dots n_{out}$. Then the L-layers Kolmogorov-Arnold Networks can be defined by:
\begin{equation}
    \mathrm{KAN} (\vx) = \left(\mathbf{\Phi}_{L-1} \circ \mathbf{\Phi}_{L-2} \circ \cdots \circ \mathbf{\Phi}_1 \circ \mathbf{\Phi}_0\right) \vx, \quad \vx\in \mathbb{R}^d.
\end{equation}
In vanilla KAN \cite{liu2024kan}, every univariate function $\phi$ is approximated via a summation with cubic spline:
\begin{equation}
    \phi_{\mathrm{spline}}\left(x\right)=w_b \mathrm{silu}\left(x\right)+w_s \left(\sum_i c_i B_i(x)\right),
\end{equation}
where $c_i, w_b, w_s$ are trainable parameters, and $B_i$ is the spline. Intuitively, to replace cubic interpolation with Sinc, we can define:

\begin{equation}\label{eq: old_sinckan}
    \phi_{\mathrm{single}}\left(x\right) = \sum_{i=-N}^N c_{i} S(i, h)(x),
\end{equation}
where $c_i$ is trainable parameters, if $h$ is set to the optimal value \cref{optimal_h}, the optimal approximation of $f(x)$ in \cref{eq: sinc_approx} is $c_i^*=f(ih) , \forall i=-N,\cdots,N$. However, to replace the interpolation method successfully, the aforementioned techniques require further investigations:

\paragraph{Optimal $h$.}
    
As we discussed in \cref{section: sinc_numerical_methods}, it is impractical to set a single optimal $h$ in machine learning frameworks. Thus in SincKAN, we propose an extension to the Sinc approximation with a mixture of different step sizes $h_j$:
\begin{equation}
    \phi_{\mathrm{multi}}\left(x\right) = \sum_{j=1}^{M}\sum_{i=-N}^N c_{i,j} S(i, h_j)(x),
\end{equation}
where $c_{i,j}$ are trainable parameters. \cite{stenger2012numerical} states that, if the chosen $h$ is larger than the optimal value predicted by \cref{optimal_h}, the interpolation is less accurate near the origin and more accurate farther away from the origin; if the chosen $h$ is smaller than the optimal value predicted by \cref{optimal_h}, the interpolation is more accurate near the origin and less accurate farther away from the origin. Herein, combining different $h$ with adaptive weights can result in a more accurate approximation than the optimal $h$ and doesn't need to calculate the optimal $h$. Thus, compared to \cref{eq: old_sinckan}, expanding the approximation by a summation of several different $h_j$ can not only avoid determining $h$ for every specific function but also improve the accuracy.

\paragraph{Coordinate transformation.}

Another challenge is the choice of coordinate transformation which is also problem-specific \cite{stenger2000summary}. Let's inherent the notation of \cref{section: sinc_numerical_methods}, and suppose $\mathcal{X}=\{x_i\}_{i=1}^{N}$ is the ordered set of input points with $x_1 \leq x_2 \leq \cdots \leq x_N$, then we can define the open interval $(a,b)$ by $a=x_1-\epsilon, b=x_N+\epsilon$, where $\epsilon$ is a chosen number, and $\xi_1=\psi^{-1}(x_1),\xi_N=\psi^{-1}(x_N)$. Thus, the interval of input points changes to $\left[\xi_1,\xi_N\right]$ from $\left[x_1,x_N\right]$. However, if we perform such a transformation for every sub-layer, the scale of the input becomes larger and inconsistent, making the network converge slower \cite{ioffe2015batch}. Herein, we argue that the normalization for the input of every layer is necessary, and \cite{bozorgasl2024wav} already utilizes the batch normalization \cite{ioffe2015batch} on every layer to enhance the performance of KANs. In our SincKAN, a normalizing transformation, $\phi (x)=\frac{x-\mu}{\sigma}$ is introduced, where $\sigma$ is the scaling factor and $\mu$ is the shifting factor. Composing $\phi$ and $\psi$ still meets the condition of $\psi$ in \cref{Theorem: transformation} and the transformed function $f(\psi \circ \phi^{-1}(\xi))$ also satisfies assumptions 1 and 2 in \cref{Theorem: Stenger} with $\alpha^{\star}, \beta^\star$ and $d^\star$. Consequently, the optimal value of the step size $h$ is changed to $\sqrt{\pi d^\star/\beta^\star N}$. 

Let us define the normalized coordinate transformation $\gamma^{-1} (x) := \phi \circ \psi^{-1}(x)$ such that $\gamma^{-1}: \left(a,b\right) \rightarrow \left(-\infty,\infty\right)$ and $\left[x_1,x_N\right] \rightarrow \left[\frac{\xi_1-\mu}{\sigma},\frac{\xi_N-\mu}{\sigma}\right]$, where $\sigma,\mu$ satisfies $\left[\frac{\xi_1-\mu}{\sigma},\frac{\xi_N-\mu}{\sigma}\right] \subset \left[-1,1\right]$. Herein, instead of coordinate transformation $\psi$, we use normalized coordinate transformation $\gamma$  in SincKAN. As for the changing of optimal $h$ with different $\sigma,\mu$, the summation of different $h_j$ implemented in SincKAN makes it easy to meet the fixed scale $\left[-1,1\right]$ \textit{i.e.} we can depend the set $\{h_j\}$ on the domain $\left[-1,1\right]$ regardless of $\sigma,\mu$. 

3) \textbf{Exponential decay}

In \cref{Theorem: transformation}, $f$ should satisfy the condition of exponential decay which constrains that $f(-\infty)=f(+\infty)=0$. To utilize the Sinc methods on general functions, \cite{richardson2011sinc} interpolates the subtraction $g-f$ instead of $f$, where $g$ is the linear function that has the same value as $f$ at the endpoints. In our SincKAN, we introduce a learnable linear function as a skip-connection to approximate the subtraction. 

Finally, combining the three aforementioned approaches, we can define our learnable activation function in SincKAN:
\begin{equation}\label{eq: Sinc activation function}
    \phi_{\mathrm{sinc}}\left(x\right) = c_1x+c_2+ \sum_{j=1}^{M}\sum_{i=-N}^N c_{i,j} S(i, h_j)(\gamma^{-1}(x)),
\end{equation}
where $c_1, c_2, c_{i,j}$ are the learnable parameters $S$ is the Sinc function and $\gamma$ is the normalized transformation.

\section{Experiments}\label{Section: experiments}
In this section, we will demonstrate the performance of SincKANs through experiments including approximating functions and solving PDEs, compared with several other representative networks: Multilayer perceptron (MLP) which is the classical and most common network used in PINNs, Modified MLP which is proposed to project the inputs to a high-dimensional feature space to enhance the hidden layers' capability, KAN which is proposed to replace MLP in AI for Science, and ChebyKAN which is proposed to improve the performance by combining KAN with the known approximation capabilities of Chebyshev polynomials and has already been examined in \cite{shukla2024comprehensive}. In this paper, we choose to implement the normalized transformation $\gamma (x)=\tanh (x)$, and the linear skip connection $w_1 \in \mathbb{R}^{n_{in}\times n_{out}},w_2 \in \mathbb{R}^{n_{out}}$. Note that, we also observed the instability of ChebyKAN highlighted in \cite{shukla2024comprehensive}, and the ChebyKAN used in our experiments is actually the modified ChebyKAN proposed by \cite{shukla2024comprehensive} which has $\tanh$ activation function between each layers. The rest details of the used networks are provided in \cref{Appendix: networks}. The other details including hyperparameters can be found in \cref{Appendix: expeirment details}. All code and data-sets accompanying this manuscript are available on GitHub at \url{https://github.com/DUCH714/SincKAN}.
\subsection{Learning for approximation}
Approximating a function by given data is the main objective of KANs with applications in identifying relevant
features, revealing modular structures, and discovering symbolic formulas \cite{liu2024kan2}. Additionally, in deep learning, the training process of a network can be regarded as approximating the map between complex functional spaces, thus the accuracy of approximation directly indicates the capability of a network. Therefore, we start with experiments on approximation to show the capability of SincKAN and verify whether SincKAN is a competitive network. In this section, to have consistent results with KAN, we inherit the metric RMSE which is used in KAN.

Sinc numerical methods are recognized theoretically and empirically as a powerful tool when dealing with singularities. However, in machine learning instead of numerical methods, we argue that SincKAN can be implemented in general cases. To demonstrate that SincKAN is robust, we conducted a series of experiments on both smooth functions and singularity functions: in \cref{table: function approxmation} the first four functions show the results of analytic functions including functions in which cubic splines interpolation is good at, and the last four functions show the results of functions with singularities which Sinc interpolation is good at. The details of the used functions can be found in \cref{Appendix: details of approximate functions}.
\begin{table}[ht]
\caption{RMSE of functions for approximation}
\label{table: function approxmation}
\begin{center}
\begin{adjustbox}{width=\columnwidth, center}
\begin{tabular}{llllll}
\multicolumn{1}{c}{\bf Function name}  & \multicolumn{1}{c}{\bf MLP} & \multicolumn{1}{c}{\bf modified MLP} &\multicolumn{1}{c}{\bf KAN} & \multicolumn{1}{c}{\bf ChebyKAN} & \multicolumn{1}{c}{\bf SincKAN (ours)} 
\\ \hline \\
\textit{sin-low} & $1.51 e\mbox{-}2 \pm 2.01 e\mbox{-}2$ & $7.29e\mbox{-}4 \pm 2.98 e\mbox{-}4$ & $1.27e\mbox{-}3 \pm 3.13e\mbox{-}4$ &
$1.76e\mbox{-}3 \pm 3.19e\mbox{-}4$ &
\textcolor{red}{$3.55 e\mbox{-}4 \pm 3.08 e\mbox{-}4$} \\ 
\textit{sin-high} & $7.07e\mbox{-}1 \pm 6.44e\mbox{-}8$ & $7.07 e\mbox{-}1 \pm 1.15 e\mbox{-}5$ & $7.06e\mbox{-}1 \pm 1.36e\mbox{-}3$ &
$5.70e\mbox{-}2 \pm 5.99 e\mbox{-}3$ &
\textcolor{red}{$3.94e\mbox{-}2 \pm 5.36e\mbox{-}3$} \\ 
\textit{bl} & $7.59e\mbox{-}4 \pm 1.13e\mbox{-}3$ & $5.73e\mbox{-}4 \pm 4.06e\mbox{-}4$ & $2.54e\mbox{-}4 \pm 7.99e\mbox{-}5$ &
$1.81e\mbox{-}3 \pm 6.98e\mbox{-}4$ &
\textcolor{red}{$4.76e\mbox{-}5 \pm 4.25e\mbox{-}5$} \\ 
\textit{double exponential} & $1.95e\mbox{-}3 \pm 8.17e\mbox{-}4$ & $7.77e\mbox{-}5 \pm 4.03e\mbox{-}5$ & $2.15e\mbox{-}4 \pm 1.52e\mbox{-}4$ &
$3.11e\mbox{-}3 \pm 2.16 e\mbox{-}3$ &
\textcolor{red}{$7.06e\mbox{-}5 \pm 1.09e\mbox{-}5$} \\ 
\textit{sqrt} & $3.06e\mbox{-}3 \pm 9.34 e\mbox{-}4$ & \textcolor{red}{$4.46 e\mbox{-}5 \pm 5.51e\mbox{-}5$} & $4.79e\mbox{-}4 \pm 1.23e\mbox{-}4$ &
$3.69e\mbox{-}3 \pm 1.27 e\mbox{-}3$ &
$3.24e\mbox{-}4 \pm 1.31e\mbox{-}4$\\ 
\textit{multi-sqrt} & $2.06e\mbox{-}3 \pm 1.16e\mbox{-}3$ & $4.59e\mbox{-}4 \pm 4.86e\mbox{-}4$ & $3.61e\mbox{-}4 \pm 8.67 e\mbox{-}5$ &
$2.34e\mbox{-}3 \pm 1.17e\mbox{-}3$ &
\textcolor{red}{$2.14e\mbox{-}4 \pm 2.49 e\mbox{-}4$} \\ 
\textit{piece\mbox{-}wise} & $2.01e\mbox{-}2 \pm 5.16e\mbox{-}3$ & $3.76e\mbox{-}2 \pm 1.83e\mbox{-}2$ & $5.84e\mbox{-}2 \pm 1.03e\mbox{-}2$ &
$7.28e\mbox{-}3 \pm 9.59 e\mbox{-}4$ &
\textcolor{red}{$2.14e\mbox{-}3 \pm 7.76e\mbox{-}4$} \\ 
\textit{spectral-bias} & $4.18e\mbox{-}3 \pm 1.18 e\mbox{-}3$ & $1.59e\mbox{-}3 \pm 2.39e\mbox{-}4$ & $4.73e\mbox{-}2 \pm 9.94 e\mbox{-}3$ &
$5.60e\mbox{-}3 \pm 2.56 e\mbox{-}4$ &
\textcolor{red}{$1.48e\mbox{-}3 \pm 1.82 e\mbox{-}4$} \\ 
\end{tabular}
\end{adjustbox}
\end{center}
\end{table}

\begin{figure}[ht]
    \centering
    \subfigure[\label{piecewise_solution}]{\includegraphics[width=0.3\linewidth]{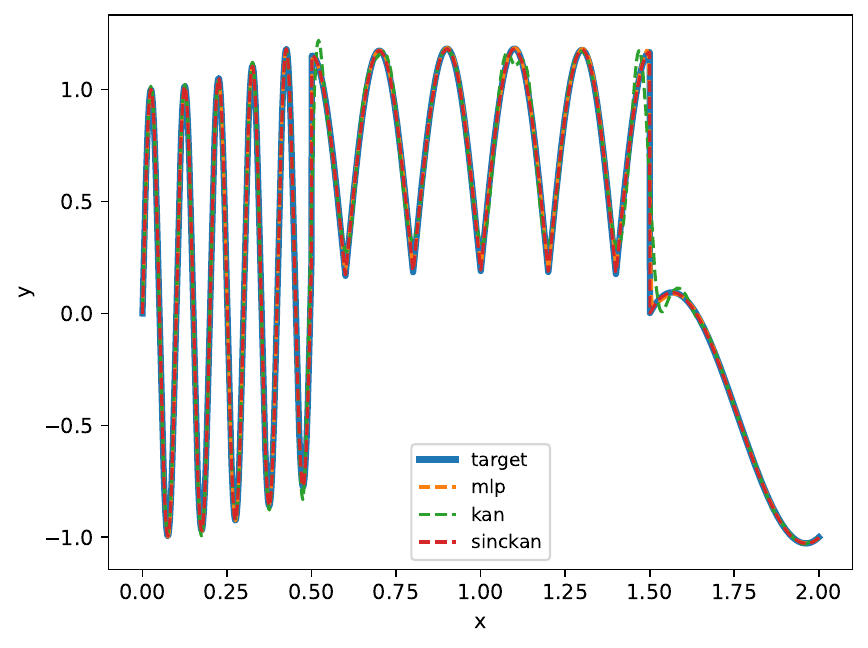}}
    \subfigure[\label{piecewise_error}]{\includegraphics[width=0.3\linewidth]{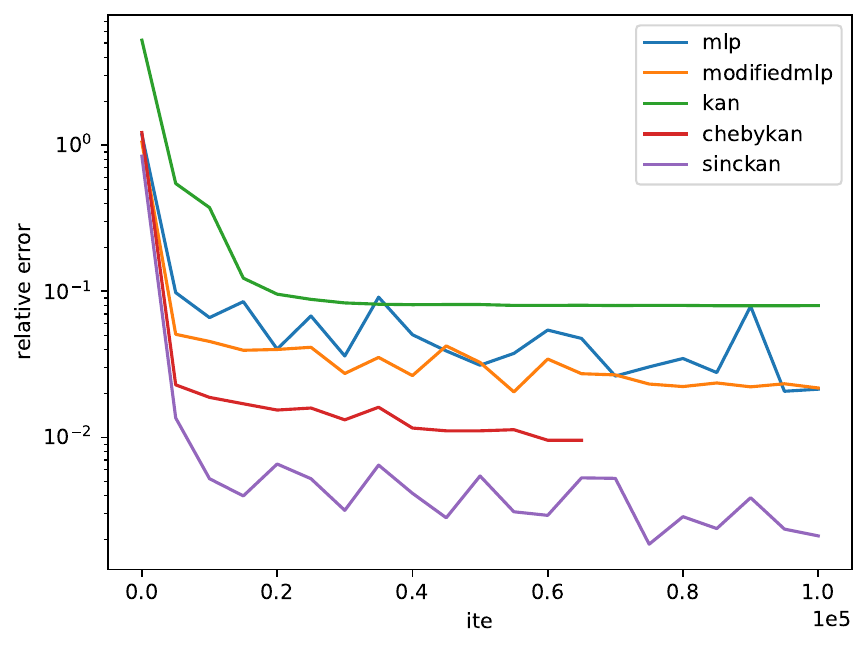}}\\
    \subfigure[\label{piecewise_sub1}]{\includegraphics[width=0.3\linewidth]{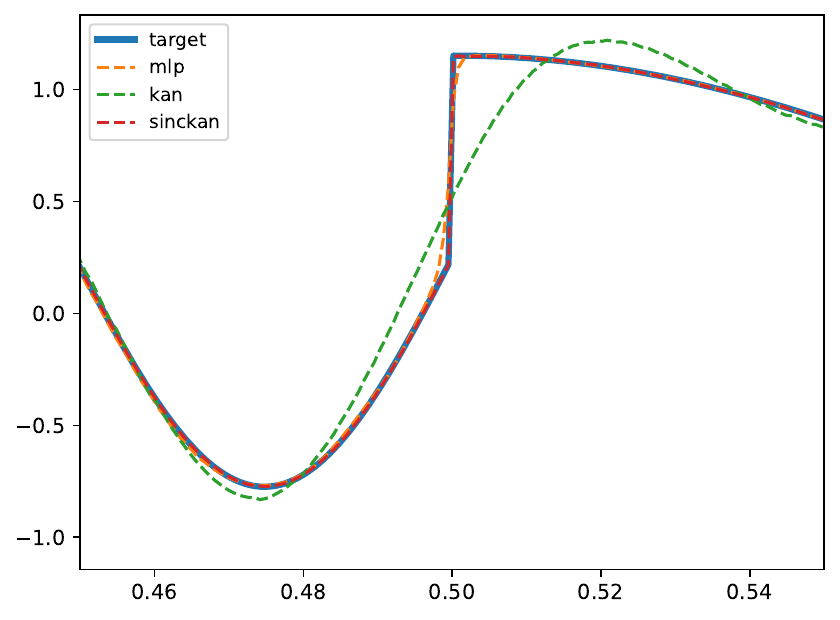}}
    \subfigure[\label{piecewise_sub2}]{\includegraphics[width=0.3\linewidth]{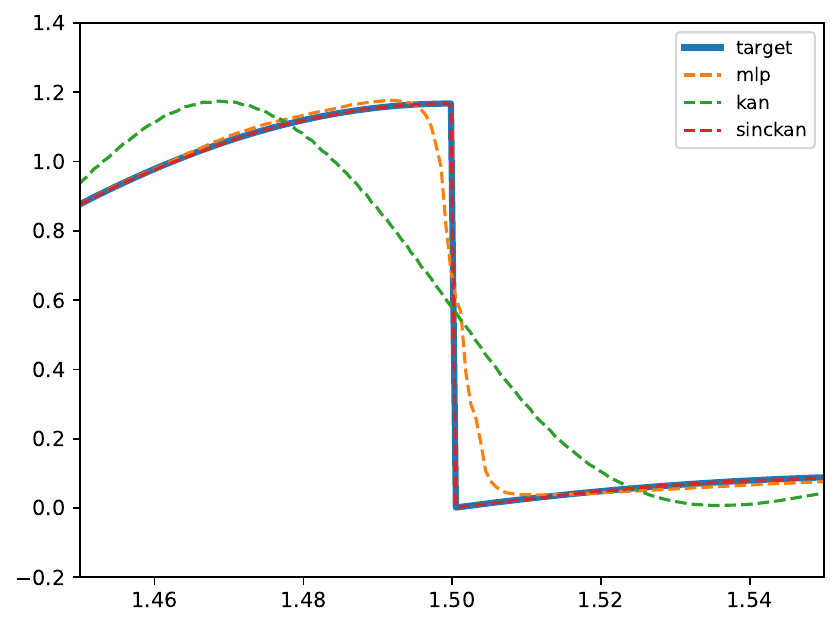}}
    \caption{\cref{piecewise_solution} depicts the function of 'piece-wise' in \cref{table: function approxmation} and compares the performance of SincKAN with MLP and KAN. \cref{piecewise_error} demonstrate the convergence of relative error for all networks, note that although the ChebyKAN we used is the modified ChebyKAN, its training is still unstable. Herein, the results of ChebyKAN in this paper are always the last valid error. \cref{piecewise_sub1} and \cref{piecewise_sub2} demonstrate the singularities in detail and show that the SincKAN can approximate the singularities well while MLP and KAN have obvious differences.}
    \label{fig:enter-label}
\end{figure}

The results in \cref{table: function approxmation} show that SincKAN achieves impressive performance on low-frequency functions (sin-low), high-frequency functions (sin-high), continuous but non-differentiable functions (multi-sqrt) and discontinuous functions (piece-wise). Furthermore, the last function (spectral-bias)  is designed to evaluate the ability to address the prevalent phenomenon of spectral bias by \cite{rahaman2019spectral}, and the corresponding result in \cref{table: function approxmation} indicates that SincKAN maximally alleviates the spectral bias. Additionally, we also evaluate every network on the finer grid to test their generalization, we put the results on \cref{Appendix: fine grids}. The comparison of cost is in \cref{Appendix: cost}. Furthermore, as an important aspect of understanding SincKANs, we plot some interior $\phi$ in \cref{Appendix: interior}.
\subsubsection{Selecting h}
Utilizing a set of $\{h_i\}$ instead of a single step size $h$ is a novel approach that we developed specifically for SincKANs. To evaluate the effectiveness of this approach, in this section, we design a comprehensive experiment. Suppose $h_{min}=\min \{h_i\}, h_{max}=\max \{h_i\}$, based on the discussion of \cref{Section: sinckan}, the ideal case is the optimal $h^*=\sqrt{\pi d^\star/\beta^\star N} \in \left(h_{min}, h_{max}\right)$. In the experiments, we  provide two types of the set:
\begin{enumerate}
    \item inverse decay $\{h_i\}_{i=1}^{M}$: $h_i=1/ih_0$,
    \item exponential decay $\{h_i\}_{i=1}^{M}$: $h_i=1/h_0^i$.
\end{enumerate}
Thus the hyperparameters of the set $\{h_i\}$ are the base number $h_0$ and the number of the set $M$. 

We train SincKAN on \textit{sin-low} and \textit{ sin-high} functions with $M=1,6,12, 24$ for inverse decay and $M=1,2,3$ for exponential decay and $h_0=2.0, \pi, 6.0, 10.0$. Besides, the number of discretized points $N_{points}$ and the degree $N_{degree}$ ($N_{degree}=2N+1$, where $N$ is the notation in \cref{eq: Sinc activation function}) also influence the performance of SincKAN for different $\{h_i\}_{i=1}^{M}$, we empirically set $N_{degree}=100$, and $N_{points}=5000$ in this experiment.

\begin{figure}[ht]
  \centering 
  \subfigure[\label{sin_low_inverse}]{\includegraphics[width=0.4\linewidth]{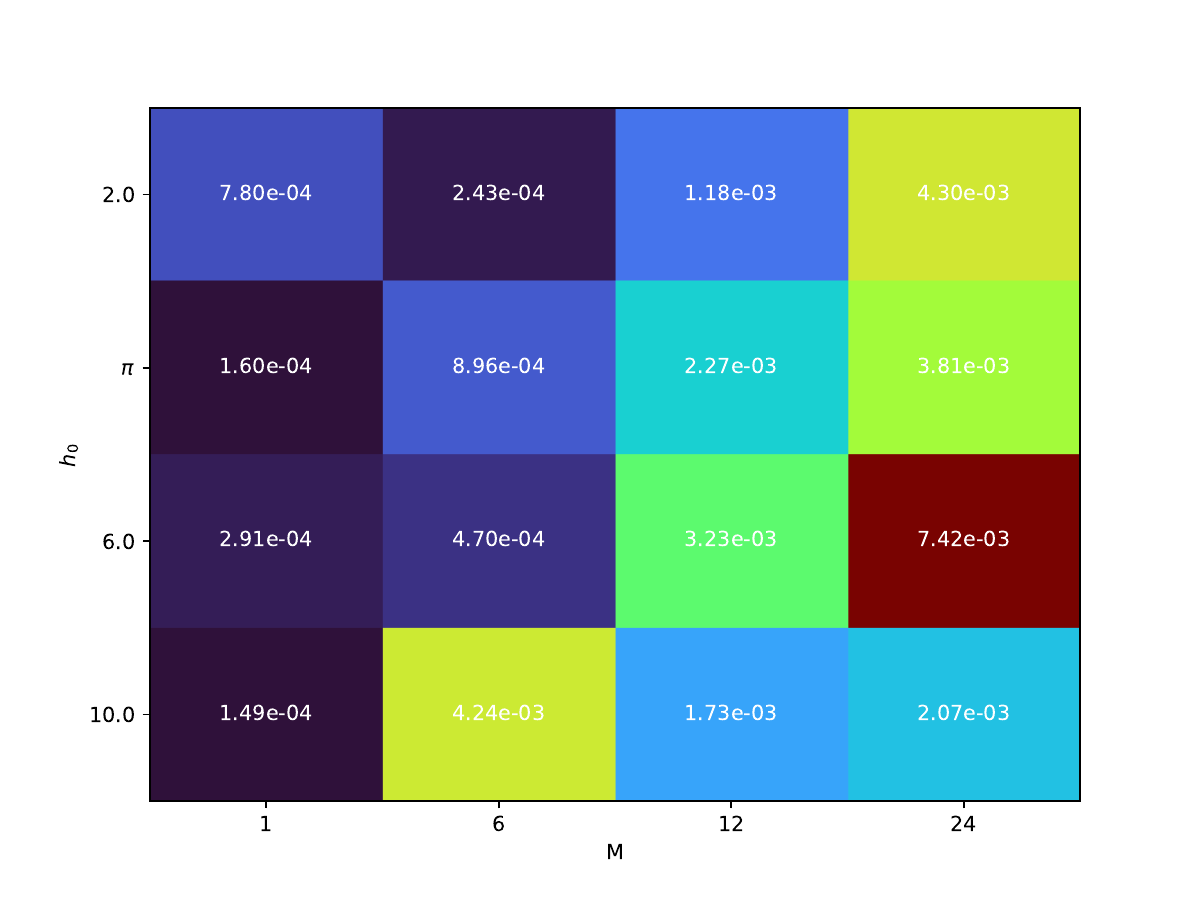}} 
  \subfigure[\label{sin_high_inverse}]{\includegraphics[width=0.4\linewidth]{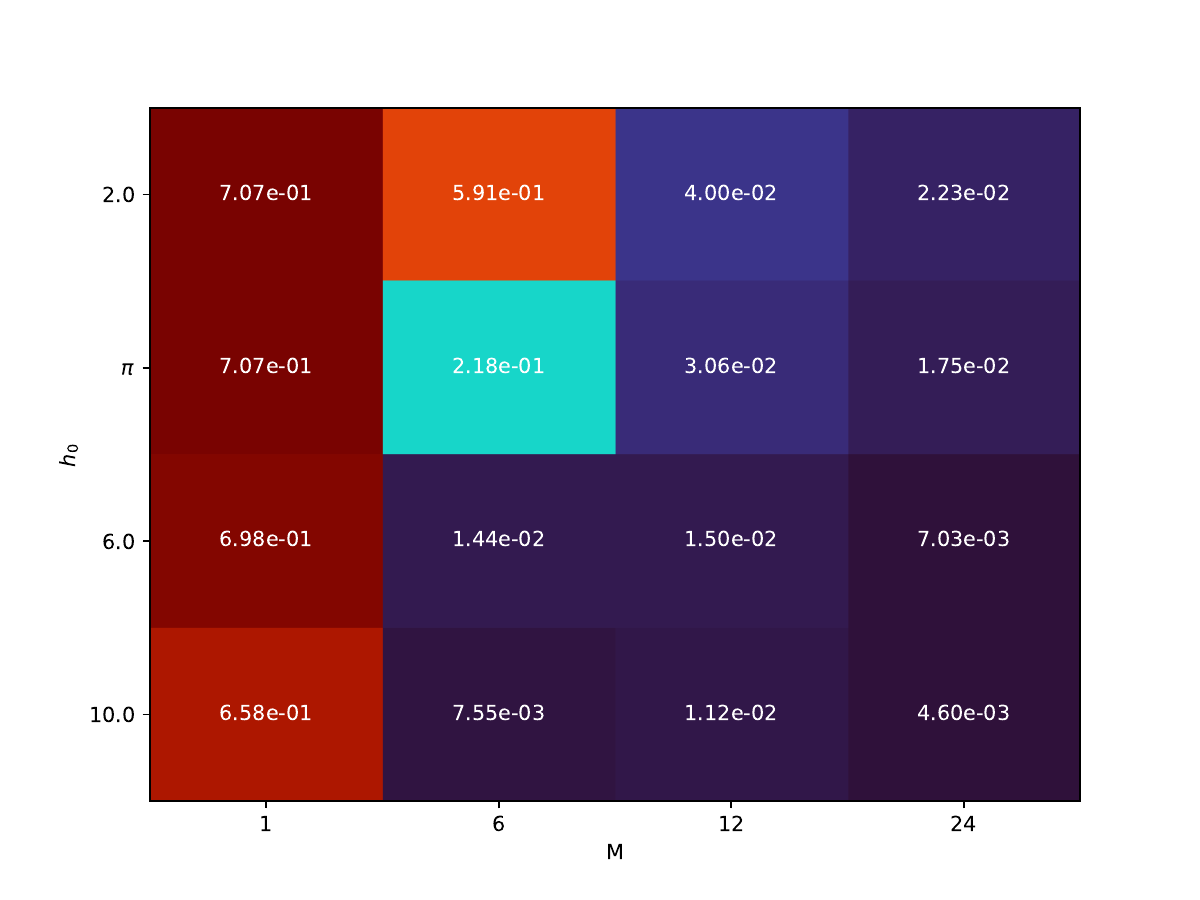}}
  \caption{\cref{sin_low_inverse} shows the inverse decay approach on sin-low function; \cref{sin_high_inverse} shows the inverse decay approach on sin-high function.}
  \label{fig:select_h_inverse}
\end{figure}

The results are illustrated in \cref{fig:select_h_inverse} for inverse decay and \cref{fig:select_h_exp} for exponential decay, and the details including the corresponding error bars are shown in \cref{Appendix: select_h}. For sin-low function, the best RMSE $1.49e\mbox{-}4 \pm 8.74e\mbox{-}5$ is observed with $h_0=10.0$ and $M=1$; for sin-high function, the best RMSE $4.60e\mbox{-}3 \pm 3.70e\mbox{-}4$ is observed in inverse decay with $h_0=10.0$ and $M=24$. 

The experiments use two divergent Fourier spectra ($4\pi$, and $400\pi$), and get extremely different optimal hyperparameters $M$. The results conclude that: to obtain an accurate result, one can use small $M$ with large $h_0$ for a low-frequency function and can use large $M$ with large $h_0$ for a high-frequency function. Although the experiments show that the SincKAN is sensitive to the approximated function, the RMSE is accurate enough for both sin-low and sin-high in inverse decay with $M=6$ and $h_0=10.0$. 

\subsubsection{Relationship between degree and size of data}
In Sinc numerical methods, the number of the sampled points is equal to the degree because each degree requires a corresponding value $f(jh)$ at the point $jh$ \textit{i.e.} $N_{degree}=N_{points}$. However, in SincKAN, $N_{degree}=N_{points}$ is impractical, and it is also not necessary because Sinc numerical methods can be regarded as a single-layer representation while our SincKAN is a multi-layer representation where the multi-layer representation has an exponentially increasing capability with depth \cite{yarotsky2017error}. To explore the relationship between degree and size of data, we train our SincKAN with different $N_{degree}$ and $N_{points}$. The results are shown in \cref{Appendix: scaling law} and reveal that it is unnecessary to maintain $N_{degree}=N_{points}$ in SincKANs.
\subsection{Learning for PIKANs}
Solving PDEs is the main part of scientific computing, and PINNs are the representative framework for solving PDEs by neural networks. In this section, we solve a series of challenging PDEs to show the performance of SincKAN. At first, we select several classical PDEs to verify the robustness of SincKAN, the results are shown in \cref{table:pikan_pde}, and the details of the PDEs can be found in \cref{Appendix: details of pde}.
\begin{table}[ht]
\caption{Relative L2 error for chosen PDE problems}
\label{table:pikan_pde}
\begin{center}
\begin{adjustbox}{width=\columnwidth, center}
\begin{tabular}{llllll}
\multicolumn{1}{c}{\bf Experiments}  & \multicolumn{1}{c}{\bf MLP} & \multicolumn{1}{c}{\bf modified MLP} &\multicolumn{1}{c}{\bf KAN} & \multicolumn{1}{c}{\bf ChebyKAN} & \multicolumn{1}{c}{\bf SincKAN (ours)} 
\\ \hline \\ 
\textit{perturbed} & $2.89e\mbox{-}2\pm 3.09e\mbox{-}2$ & $6.30e\mbox{-}1\pm1.14e\mbox{-}1$ &  $4.48e\mbox{-}3 \pm 4.20e\mbox{-}3 $& $6.73e\mbox{-}1 \pm 1.02e\mbox{-}1 $& \color{red}{$1.88e\mbox{-}3 \pm 8.55e\mbox{-}4$} \\ 
\textit{nonlinear} & $3.92e\mbox{-}1 \pm 2.36e\mbox{-}5$ & $1.56e\mbox{-}2 \pm 2.10e\mbox{-}2$ & \color{red}{$6.15e\mbox{-}4 \pm 7.96e\mbox{-}4$} & $7.78e\mbox{-}1 \pm 2.67e\mbox{-}2$  & $1.77e\mbox{-}3 \pm 1.06e\mbox{-}3$ \\ 
\textit{bl-2d} & $2.38e\mbox{-}1\pm6.22e\mbox{-}2$ & $5.34e\mbox{-}2 \pm 1.91e\mbox{-}2$ &$1.19e\mbox{-}2 \pm 4.22e\mbox{-}3$ & $5.97e\mbox{-}2\pm 3.83e\mbox{-}2$ & \color{red}{$2.31e\mbox{-}3\pm7.10e\mbox{-}4$} \\
\textit{ns-tg-u} & $8.14e\mbox{-}5 \pm 1.96e\mbox{-}6$ & \color{red}{$2.14e\mbox{-}5 \pm 2.67e\mbox{-}6$} & $3.21e\mbox{-}4 \pm 2.02e\mbox{-}5$ & $6.43e\mbox{-}2 \pm 2.70e\mbox{-}2$& $6.51e\mbox{-}4 \pm 7.03e\mbox{-}5$\\      
\textit{ns-tg-v} & $8.30e\mbox{-}5 \pm 2.47e\mbox{-}6$ & \color{red}{$1.91e\mbox{-}5 \pm 1.63e\mbox{-}6$} & $4.04e\mbox{-}4 \pm 1.25e\mbox{-}4$ & $5.86e\mbox{-}2 \pm 4.15e\mbox{-}2$& $1.34e\mbox{-}3 \pm 4.38e\mbox{-}4$
\end{tabular}
\end{adjustbox}
\end{center}
\end{table}
\subsubsection{Boundary layer problem}
To intuitively show the performance of SincKAN compared with other networks, we conducted additional experiments on the boundary layer problem:
\begin{equation} \label{eq: bl}
u_{xx}/\epsilon+u_{x}=0, x \in [0,1]
\end{equation}
with the exact solution $u(x)=\exp (-\epsilon x)$. As $\epsilon$ increases, the width of the boundary layer (left) decreases, and the complexity of learning increases. The results shown in Table 3 and Fig. 4 reveal that SincKANs can handle the boundary layer effectively, while other networks struggle when $\epsilon$ is large.
\begin{table}[ht]
\caption{Relative L2 error for different $\epsilon$ in \cref{eq: bl}}
\label{table:bl_pikan}
\begin{center}
\begin{adjustbox}{width=\columnwidth, center}
\begin{tabular}{llllll}
\multicolumn{1}{c}{\bf $\epsilon$ } & \multicolumn{1}{c}{\bf MLP} & \multicolumn{1}{c}{\bf Modified MLP} &\multicolumn{1}{c}{\bf KAN} &\multicolumn{1}{c}{\bf ChebyKAN} & \multicolumn{1}{c}{\bf SincKAN (ours)} 
\\ \hline \\
1    & $ 6.60e\mbox{-}5 \pm 1.91e\mbox{-}5 $ &$ 3.88e\mbox{-}6 \pm 7.22e\mbox{-}7 $   & $ 5.97e\mbox{-}6 \pm 5.24e\mbox{-}6 $ & \color{red}{$ 1.98e\mbox{-}6 \pm 4.51e\mbox{-}7 $} & $ 7.78e\mbox{-}5 \pm 1.14e\mbox{-}4 $ \\
10   & $ 2.83e\mbox{-}4 \pm 4.22e\mbox{-}5 $& $ 1.69e\mbox{-}4 \pm 6.85e\mbox{-}5 $ &$ 3.23e\mbox{-}5 \pm 1.81e\mbox{-}5 $ & \color{red}{$4.45e\mbox{-}6 \pm 4.01e\mbox{-}7$} & $1.14e\mbox{-}4 \pm 1.64e\mbox{-}4 $\\
100  & $ 1.29e\mbox{-}3 \pm 3.24e\mbox{-}4 $& $ 6.25e\mbox{-}4 \pm 2.27e\mbox{-}4 $ & $ 1.25e\mbox{-}2 \pm 2.62e\mbox{-}3 $ & $ 5.27e\mbox{-}4 \pm 6.55e\mbox{-}4 $& \color{red}{$1.68e\mbox{-}4 \pm 6.16e\mbox{-}5$} \\
1000 &$ 9.87 \pm 8.70 $ & $ 1.53e\mbox{-}1 \pm 5.59e\mbox{-}2 $    & $ 11.3 \pm 8.79 $ & $ 10.9 \pm 7.18 $ & \color{red}{$5.48e\mbox{-}3 \pm 3.45e\mbox{-}3$}
\end{tabular}
\end{adjustbox}
\end{center}
\end{table}

\begin{figure}[ht]
  \centering 
  \subfigure[\label{bl_solution_sinckan}]{\includegraphics[width=0.329\linewidth]{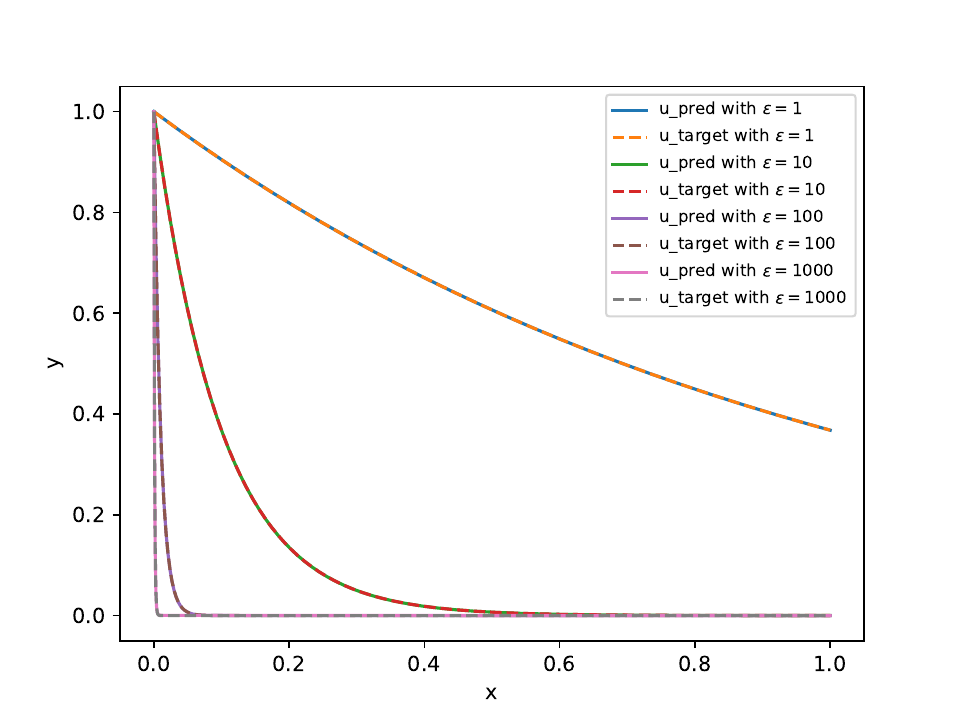}} 
  \subfigure[\label{bl_loss_sinckan}]{\includegraphics[width=0.329\linewidth]{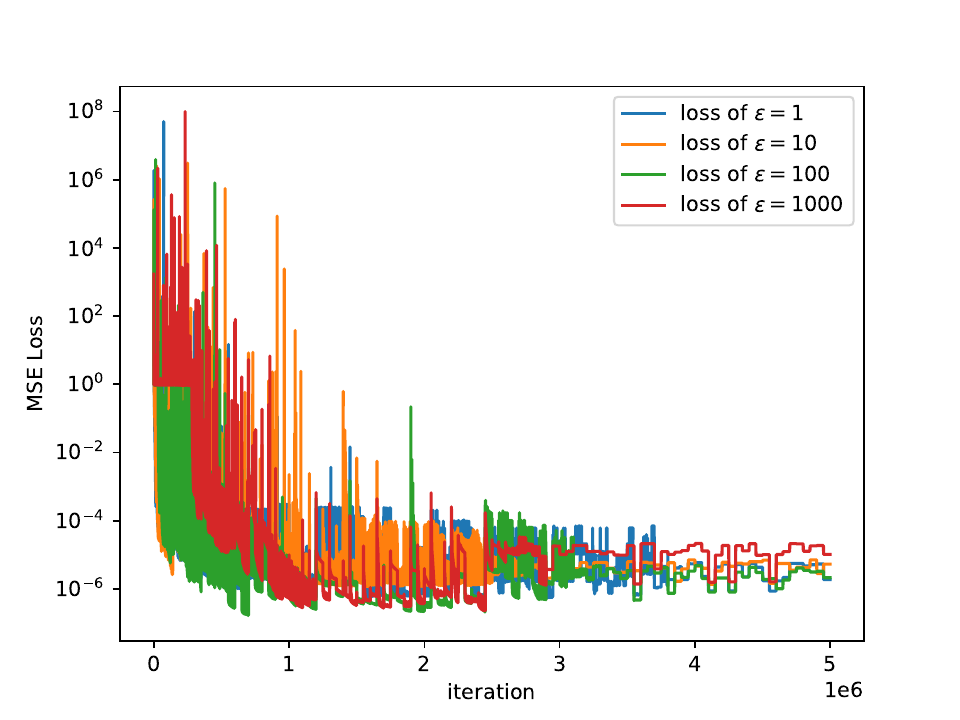}}
  \subfigure[\label{bl_1000_sinckan}]{\includegraphics[width=0.329\linewidth]{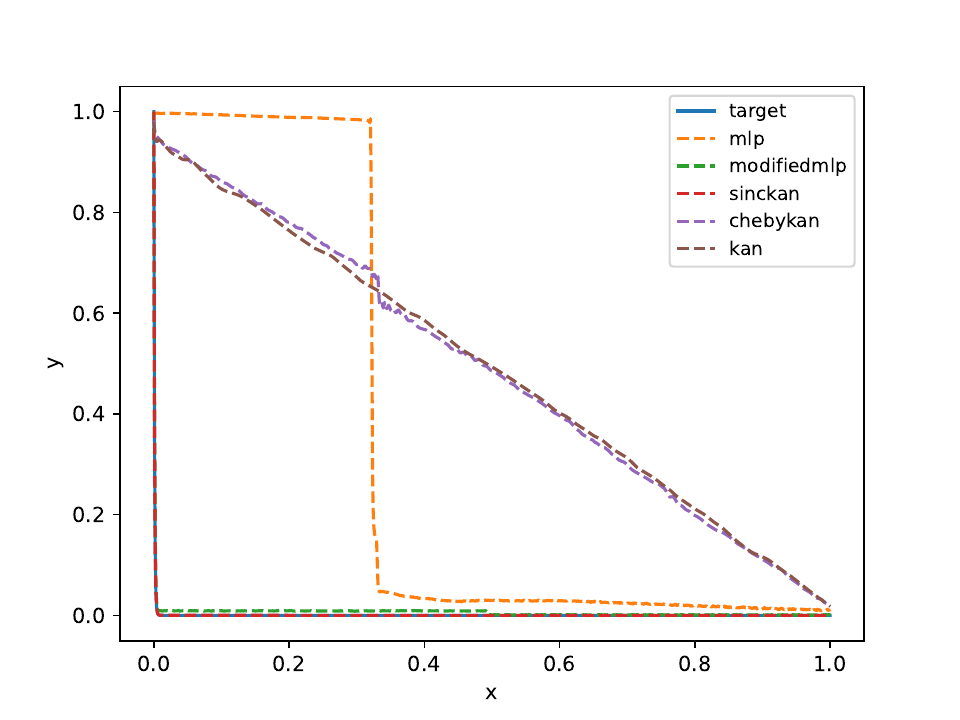}}
  \caption{Demonstration of boundary layer problems, \cref{bl_solution_sinckan} depicts the exact solution of \cref{eq: bl} with different $\epsilon$ and the corresponding predicted solution by SincKAN, states that SincKAN can solve \cref{eq: bl} properly even with an extremely narrow boundary layer. \cref{bl_loss_sinckan} depicts the convergence of the training loss function of SincKAN with different $\epsilon$. \cref{bl_1000_sinckan} demonstrates the results of different networks when solving \cref{eq: bl} with $\epsilon=1000$ and reveals that the derivative in the boundary layer is so large that other networks cannot approximate the boundary layer well.}
  \label{fig:bl_sinckan}
\end{figure}	

\subsubsection{Ablation study}
Compared with Sinc numerical methods, SincKANs have a normalized transformation; compared to KANs, SincKANs have a skip connection with linear functions. However, the Sinc numerical methods also have some choices of coordinate transformations and KANs also have a skip connection with SiLU functions. In this section, we conduct an ablation study on SincKANs with non-normalized transformation and the SiLU skip connection to verify the effect of the two proposed modules. This experiment uses Burger's equation \cref{eq:burgers} and time-dependent nonlinear equation \cref{eq:tnonlinear}.
\cref{table:ablation} shows the results of the ablation study where $\psi(x)=\log(\frac{x-a}{b-x})$, $\gamma(x)=\tanh(x)$, $\mathrm {Linear}(x)=w_1x+w_2+\phi(x)$, and $\mathrm {SiLU}(x)=w_b\mathrm{silu}(x)+w_s\phi(x)$. The non-normalized transformation performs poorly, even compared to the cases without transformations. Although the linear skip connection is not the best for both equations, it is the most stable approach for SincKANs.
\begin{table}[ht]
\caption{Relative L2 error for ablation study}
\label{table:ablation}
\begin{center}
\begin{tabular}{ccccll}
\multicolumn{1}{c} {$\psi$}  & \multicolumn{1}{c}{$\gamma$} & \multicolumn{1}{c}{\bf Linear }  &\multicolumn{1}{c}{\bf SiLU} & \multicolumn{1}{c}{\bf T-nonlinear} & \multicolumn{1}{c}{\bf Burgers' equation}
\\ \hline \\ 
\ding{56} & \ding{56} & \ding{56} & \ding{56} & $9.20e\mbox{-}4\pm4.31e\mbox{-}4$ & $1.57e\mbox{-}2\pm4.31e\mbox{-}3$\\
\ding{56} & \ding{52} & \ding{56} & \ding{56} &  $5.80e\mbox{-}4\pm1.89e\mbox{-}4$ &\color{red}{$6.21e\mbox{-}4\pm1.96e\mbox{-}4$}\\
\ding{56} & \ding{52} & \ding{52} & \ding{56} & $2.44e\mbox{-}4\pm6.08e\mbox{-}5$ & $3.12e\mbox{-}3\pm2.48e\mbox{-}3$ \\
\ding{56} & \ding{52} & \ding{56} & \ding{52} & \color{red}{$1.60e\mbox{-}4\pm3.17e\mbox{-}5$} & $8.90e\mbox{-}3\pm6.76e\mbox{-}3$ \\
\ding{52} & \ding{56} & \ding{52} & \ding{56} & $1.11e\mbox{-}2\pm1.17e\mbox{-}3$& $7.36e\mbox{-}2\pm2.02e\mbox{-}2$ \\
\ding{52} & \ding{56} & \ding{56} & \ding{52} & $1.30e\mbox{-}2\pm4.00e\mbox{-}4$ & $1.12e\mbox{-}1\pm6.80e\mbox{-}3$ \\
\end{tabular}
\end{center}
\end{table}

\section{Conclusion}\label{Section: conclusion}
In this paper, we propose a novel network called Sinc Kolmogorov-Arnold Networks (SincKANs). Inspired by KANs, SincKANs leverage the Sinc functions to interpolate the activation function and successfully inherit the capability of handling singularities. To set the optimal $h$, we propose the multi-$h$ interpolation, and the corresponding experiments indicate that this novel approach is the main reason for SincKANs' superior ability in approximating complex smooth functions; to choose a proper coordinate transformation for machine learning, we propose the normalized transformation which prevents slow convergence.; to satisfy the decay condition, we introduce the skip-connection with learnable linear functions. After tackling the aforementioned challenges, SincKANs become a competitive network that can replace the current networks used for PINNs.

We begin with training on approximation problems to demonstrate the capability of SincKANs. The results reveal that SincKANs excel in most experiments with other networks. However, directly approximating the target function is an impractical objective for almost all machine learning tasks. After verifying the capability, we turn to solving PDEs in the PINNs framework. Although the SincKANs achieve impressive performance in approximation tasks for solving all chosen PDEs, SincKANs merely have the best accuracy on boundary layer problems, due to the oscillations caused by the inaccuracy of derivatives.
\paragraph{Limitations}
Approximating derivative by Sinc numerical methods is always inaccurate in the neighborhood of the Sinc end-points. To address this problem,  \cite{stenger2009polynomial} suggested using Lagrange polynomial to approximate the derivative instead of straightforwardly calculating the derivative of Sinc polynomials, \cite{wu2006sinc} used several discrete functions to replace the derivative of Sinc polynomials, etc. Unfortunately, to the best of our knowledge, there isn't an approach that can be implemented in our SincKAN when we demand the derivatives of SincKAN in PIKANs. Herein, to alleviate the inaccuracy, we choose small $h_0$, small $M$, and small $N$ so that SincKAN can solve PDEs, otherwise, the solution will have oscillations (see \cref{Appendix: oscillation}). Such kind of setting limits the capability of SincKAN and we argue that this is the main reason that SincKAN can obtain good results but not the best results for some cases. Furthermore, the inaccuracy limits SincKAN in solving high-order problems such as Korteweg–De Vries equations, and Kuramoto–Sivashinsky equations.

\paragraph{Future}
As the accuracy of approximating the derivative decreases with the order of derivative increases if the PDE merely requires the first derivatives, then the SincKANs will release the limitation to have larger enough $h_0$, $M$, and $N$ and improve the performance. In literature, to avoid calculating the high-order derivatives, MIM~\cite{lyu2022mim,li2024priori} is proposed to use the mixed residual method which transforms a high-order PDE into a first-order PDE system. SincKANs can implement this approach to calculate several first-order derivatives instead of the high-order derivatives so that SincKANs can have accurate estimations for the residual loss. Furthermore, replacing the automatic differentiation \cite{cen2024deep,yu2024fourier} by other operators is also an expected research. 
\subsubsection*{Acknowledgments}
T. Yu is supported by China Scholarship Council No. 202308090008. J. Yang is supported by the National Natural Science Foundation of China (NSFC) Grant No. 12271240, the NSFC/Hong Kong RGC Joint Research Scheme (NSFC/RGC 11961160718), and the Shenzhen Natural Science Fund (No. RCJC20210609103819018). We thank Vladimir Fanakov for fruitful discussion and constructive suggestions. 

\bibliography{iclr2025_conference}

@article{raissi2019physics,
  title={Physics-informed neural networks: A deep learning framework for solving forward and inverse problems involving nonlinear partial differential equations},
  author={Raissi, Maziar and Perdikaris, Paris and Karniadakis, George E},
  journal={Journal of Computational physics},
  volume={378},
  pages={686--707},
  year={2019},
  publisher={Elsevier}
}

@article{wojtowytsch2020can,
  title={Can shallow neural networks beat the curse of dimensionality? a mean field training perspective},
  author={Wojtowytsch, Stephan and Weinan, E},
  journal={IEEE Transactions on Artificial Intelligence},
  volume={1},
  number={2},
  pages={121--129},
  year={2020},
  publisher={IEEE}
}

@article{han2018solving,
  title={Solving high-dimensional partial differential equations using deep learning},
  author={Han, Jiequn and Jentzen, Arnulf and E, Weinan},
  journal={Proceedings of the National Academy of Sciences},
  volume={115},
  number={34},
  pages={8505--8510},
  year={2018},
  publisher={National Acad Sciences}
}

@article{karniadakis2021physics,
  title={Physics-informed machine learning},
  author={Karniadakis, George Em and Kevrekidis, Ioannis G and Lu, Lu and Perdikaris, Paris and Wang, Sifan and Yang, Liu},
  journal={Nature Reviews Physics},
  volume={3},
  number={6},
  pages={422--440},
  year={2021},
  publisher={Nature Publishing Group}
}

@article{sliwinski2023mean,
  title={Mean flow reconstruction of unsteady flows using physics-informed neural networks},
  author={Sliwinski, Lukasz and Rigas, Georgios},
  journal={Data-Centric Engineering},
  volume={4},
  pages={e4},
  year={2023},
  publisher={Cambridge University Press}
}

@article{jin2021nsfnets,
  title={{NSFnets (Navier-Stokes flow nets)}: Physics-informed neural networks for the incompressible {Navier}-{Stokes} equations},
  author={Jin, Xiaowei and Cai, Shengze and Li, Hui and Karniadakis, George Em},
  journal={Journal of Computational Physics},
  volume={426},
  pages={109951},
  year={2021},
  publisher={Elsevier}
}

@article{fang2019physics,
  title={A physics-informed neural network framework for {PDEs on 3D} surfaces: Time independent problems},
  author={Fang, Zhiwei and Zhan, Justin},
  journal={IEEE Access},
  volume={8},
  pages={26328--26335},
  year={2019},
  publisher={IEEE}
}

@article{xu2019frequency,
  title={Frequency principle: {Fourier} analysis sheds light on deep neural networks},
  author={Xu, Zhi-Qin John and Zhang, Yaoyu and Luo, Tao and Xiao, Yanyang and Ma, Zheng},
  journal={arXiv preprint arXiv:1901.06523},
  year={2019}
}

@software{jax2018github,
  author = {James Bradbury and Roy Frostig and Peter Hawkins and Matthew James Johnson and Chris Leary and Dougal Maclaurin and George Necula and Adam Paszke and Jake Vander{P}las and Skye Wanderman-{M}ilne and Qiao Zhang},
  title = {{JAX}: composable transformations of {P}ython+{N}um{P}y programs},
  url = {http://github.com/google/jax},
  version = {0.3.13},
  year = {2018},
}

@article{kingma2014adam,
  title={Adam: A method for stochastic optimization},
  author={Kingma, Diederik P and Ba, Jimmy},
  journal={arXiv preprint arXiv:1412.6980},
  year={2014}
}

@article{wang2022and,
  title={When and why {PINNs} fail to train: A neural tangent kernel perspective},
  author={Wang, Sifan and Yu, Xinling and Perdikaris, Paris},
  journal={Journal of Computational Physics},
  volume={449},
  pages={110768},
  year={2022},
  publisher={Elsevier}
}

@article{raissi2020hidden,
  title={Hidden fluid mechanics: Learning velocity and pressure fields from flow visualizations},
  author={Raissi, Maziar and Yazdani, Alireza and Karniadakis, George Em},
  journal={Science},
  volume={367},
  number={6481},
  pages={1026--1030},
  year={2020},
  publisher={American Association for the Advancement of Science}
}

@inproceedings{jin2022physics,
  title={Physics-informed neural networks for quantum eigenvalue problems},
  author={Jin, Henry and Mattheakis, Marios and Protopapas, Pavlos},
  booktitle={2022 International Joint Conference on Neural Networks (IJCNN)},
  pages={1--8},
  year={2022},
  organization={IEEE}
}

@article{yazdani2020systems,
  title={Systems biology informed deep learning for inferring parameters and hidden dynamics},
  author={Yazdani, Alireza and Lu, Lu and Raissi, Maziar and Karniadakis, George Em},
  journal={PLoS computational biology},
  volume={16},
  number={11},
  pages={e1007575},
  year={2020},
  publisher={Public Library of Science San Francisco, CA USA}
}

@article{nellikkath2022physics,
  title={Physics-informed neural networks for ac optimal power flow},
  author={Nellikkath, Rahul and Chatzivasileiadis, Spyros},
  journal={Electric Power Systems Research},
  volume={212},
  pages={108412},
  year={2022},
  publisher={Elsevier}
}

@article{cen2024deep,
  title={Deep Finite Volume Method for Partial Differential Equations},
  author={Cen, Jianhuan and Zou, Qingsong},
  journal={Journal of Computational Physics},
  pages={113307},
  year={2024},
  publisher={Elsevier}
}

@article{wang2021understanding,
  title={Understanding and mitigating gradient flow pathologies in physics-informed neural networks},
  author={Wang, Sifan and Teng, Yujun and Perdikaris, Paris},
  journal={SIAM Journal on Scientific Computing},
  volume={43},
  number={5},
  pages={A3055--A3081},
  year={2021},
  publisher={SIAM}
}

@article{liu2024kan,
  title={Kan: {Kolmogorov-Arnold} networks},
  author={Liu, Ziming and Wang, Yixuan and Vaidya, Sachin and Ruehle, Fabian and Halverson, James and Solja{\v{c}}i{\'c}, Marin and Hou, Thomas Y and Tegmark, Max},
  journal={arXiv preprint arXiv:2404.19756},
  year={2024}
}

@book{stenger2012numerical,
  title={Numerical methods based on sinc and analytic functions},
  author={Stenger, Frank},
  volume={20},
  year={2012},
  publisher={Springer Science \& Business Media}
}

@article{bozorgasl2024wav,
  title={Wav-KAN: Wavelet {Kolmogorov-Arnold} Networks},
  author={Bozorgasl, Zavareh and Chen, Hao},
  journal={arXiv e-prints},
  pages={arXiv--2405},
  year={2024}
}

@article{seydi2024unveiling,
  title={Unveiling the Power of Wavelets: A Wavelet-based {Kolmogorov-Arnold} Network for Hyperspectral Image Classification},
  author={Seydi, Seyd Teymoor},
  journal={arXiv preprint arXiv:2406.07869},
  year={2024}
}

@article{xu2024fourierkan,
  title={FourierKAN-GCF: Fourier {Kolmogorov-Arnold} Network--An Effective and Efficient Feature Transformation for Graph Collaborative Filtering},
  author={Xu, Jinfeng and Chen, Zheyu and Li, Jinze and Yang, Shuo and Wang, Wei and Hu, Xiping and Ngai, Edith C-H},
  journal={arXiv preprint arXiv:2406.01034},
  year={2024}
}

@article{howard2024finite,
  title={Finite basis {Kolmogorov-Arnold} networks: domain decomposition for data-driven and physics-informed problems},
  author={Howard, Amanda A and Jacob, Bruno and Murphy, Sarah H and Heinlein, Alexander and Stinis, Panos},
  journal={arXiv preprint arXiv:2406.19662},
  year={2024}
}

@article{aghaei2024fkan,
  title={fKAN: Fractional {Kolmogorov-Arnold} Networks with trainable Jacobi basis functions},
  author={Aghaei, Alireza Afzal},
  journal={arXiv preprint arXiv:2406.07456},
  year={2024}
}

@article{aghaei2024rkan,
  title={rKAN: Rational {Kolmogorov-Arnold} Networks},
  author={Aghaei, Alireza Afzal},
  journal={arXiv preprint arXiv:2406.14495},
  year={2024}
}

@article{seydi2024exploring,
  title={Exploring the Potential of Polynomial Basis Functions in {Kolmogorov-Arnold} Networks: A Comparative Study of Different Groups of Polynomials},
  author={Seydi, Seyd Teymoor},
  journal={arXiv preprint arXiv:2406.02583},
  year={2024}
}

@article{ss2024chebyshev,
  title={Chebyshev polynomial-based {Kolmogorov-Arnold} networks: An efficient architecture for nonlinear function approximation},
  author={SS, Sidharth},
  journal={arXiv preprint arXiv:2405.07200},
  year={2024}
}

@article{shukla2024comprehensive,
  title={A comprehensive and FAIR comparison between MLP and KAN representations for differential equations and operator networks},
  author={Shukla, Khemraj and Toscano, Juan Diego and Wang, Zhicheng and Zou, Zongren and Karniadakis, George Em},
  journal={arXiv preprint arXiv:2406.02917},
  year={2024}
}

@article{rigas2024adaptive,
  title={Adaptive training of grid-dependent physics-informed {Kolmogorov-Arnold} networks},
  author={Rigas, Spyros and Papachristou, Michalis and Papadopoulos, Theofilos and Anagnostopoulos, Fotios and Alexandridis, Georgios},
  journal={arXiv preprint arXiv:2407.17611},
  year={2024}
}

@article{wang2024kolmogorov,
  title={Kolmogorov {Arnold} Informed neural network: A physics-informed deep learning framework for solving PDEs based on {Kolmogorov Arnold} Networks},
  author={Wang, Yizheng and Sun, Jia and Bai, Jinshuai and Anitescu, Cosmin and Eshaghi, Mohammad Sadegh and Zhuang, Xiaoying and Rabczuk, Timon and Liu, Yinghua},
  journal={arXiv preprint arXiv:2406.11045},
  year={2024}
}

@article{sugihara2004recent,
  title={Recent developments of the Sinc numerical methods},
  author={Sugihara, Masaaki and Matsuo, Takayasu},
  journal={Journal of computational and applied mathematics},
  volume={164},
  pages={673--689},
  year={2004},
  publisher={Elsevier}
}

@book{stenger2016handbook,
  title={Handbook of Sinc numerical methods},
  author={Stenger, Frank},
  year={2016},
  publisher={CRC Press}
}

@article{mohsen2017accurate,
  title={Accurate function Sinc interpolation and derivative estimations over finite intervals},
  author={Mohsen, Adel AK},
  journal={Journal of Computational and Applied Mathematics},
  volume={324},
  pages={216--224},
  year={2017},
  publisher={Elsevier}
}

@article{richardson2011sinc,
  title={A sinc function analogue of Chebfun},
  author={Richardson, Mark and Trefethen, Lloyd N},
  journal={SIAM Journal on Scientific Computing},
  volume={33},
  number={5},
  pages={2519--2535},
  year={2011},
  publisher={SIAM}
}

@article{lippke1991analytical,
  title={Analytical solutions and Sinc function approximations in thermal conduction with nonlinear heat generation},
  author={Lippke, A},
  journal={ASME J. Heat Transf},
  volume={113},
  pages={5--11},
  year={1991},
  publisher={Citeseer}
}

@article{abdella2015solving,
  title={Solving differential equations using Sinc-collocation methods with derivative interpolations},
  author={Abdella, Kenzu},
  journal={Journal of Computational Methods in Sciences and Engineering},
  volume={15},
  number={3},
  pages={305--315},
  year={2015},
  publisher={IOS Press}
}

@article{abdella2009application,
  title={Application of the Sinc method to a dynamic elasto-plastic problem},
  author={Abdella, K and Yu, X and Kucuk, I},
  journal={Journal of Computational and Applied Mathematics},
  volume={223},
  number={2},
  pages={626--645},
  year={2009},
  publisher={Elsevier}
}

@article{stenger2009polynomial,
  title={Polynomial function and derivative approximation of Sinc data},
  author={Stenger, Frank},
  journal={Journal of Complexity},
  volume={25},
  number={3},
  pages={292--302},
  year={2009},
  publisher={Elsevier}
}

@article{stenger2000summary,
  title={Summary of Sinc numerical methods},
  author={Stenger, Frank},
  journal={Journal of Computational and Applied Mathematics},
  volume={121},
  number={1-2},
  pages={379--420},
  year={2000},
  publisher={Elsevier}
}

@article{ioffe2015batch,
  title={Batch normalization: Accelerating deep network training by reducing internal covariate shift},
  author={Ioffe, Sergey},
  journal={arXiv preprint arXiv:1502.03167},
  year={2015}
}

@article{fanaskov2024astral,
  title={Astral: training physics-informed neural networks with error majorants},
  author={Fanaskov, Vladimir and Yu, Tianchi and Rudikov, Alexander and Oseledets, Ivan},
  journal={arXiv preprint arXiv:2406.02645},
  year={2024}
}

@article{yu2024fourier,
  title={Fourier Spectral Physics Informed Neural Network: An Efficient and Low-Memory PINN},
  author={Yu, Tianchi and Qi, Yiming and Oseledets, Ivan and Chen, Shiyi},
  journal={arXiv preprint arXiv:2408.16414},
  year={2024}
}

@inproceedings{yao2023multiadam,
  title={Multiadam: Parameter-wise scale-invariant optimizer for multiscale training of physics-informed neural networks},
  author={Yao, Jiachen and Su, Chang and Hao, Zhongkai and Liu, Songming and Su, Hang and Zhu, Jun},
  booktitle={International Conference on Machine Learning},
  pages={39702--39721},
  year={2023},
  organization={PMLR}
}

@article{kashefi2022physics,
  title={Physics-informed PointNet: A deep learning solver for steady-state incompressible flows and thermal fields on multiple sets of irregular geometries},
  author={Kashefi, Ali and Mukerji, Tapan},
  journal={Journal of Computational Physics},
  volume={468},
  pages={111510},
  year={2022},
  publisher={Elsevier}
}

@book{trefethen2000spectral,
  title={Spectral methods in MATLAB},
  author={Trefethen, Lloyd N},
  year={2000},
  publisher={SIAM}
}

@article{yarotsky2017error,
  title={Error bounds for approximations with deep ReLU networks},
  author={Yarotsky, Dmitry},
  journal={Neural networks},
  volume={94},
  pages={103--114},
  year={2017},
  publisher={Elsevier}
}

@article{liu2024kan2,
  title={Kan 2.0: {Kolmogorov-Arnold} networks meet science},
  author={Liu, Ziming and Ma, Pingchuan and Wang, Yixuan and Matusik, Wojciech and Tegmark, Max},
  journal={arXiv preprint arXiv:2408.10205},
  year={2024}
}

@article{wu2006sinc,
  title={Sinc collocation method with boundary treatment for two-point boundary value problems},
  author={Wu, Xionghua and Kong, Wenbin and Li, Chen},
  journal={Journal of computational and applied mathematics},
  volume={196},
  number={1},
  pages={229--240},
  year={2006},
  publisher={Elsevier}
}

@inproceedings{rahaman2019spectral,
  title={On the spectral bias of neural networks},
  author={Rahaman, Nasim and Baratin, Aristide and Arpit, Devansh and Draxler, Felix and Lin, Min and Hamprecht, Fred and Bengio, Yoshua and Courville, Aaron},
  booktitle={International conference on machine learning},
  pages={5301--5310},
  year={2019},
  organization={PMLR}
}

@article{lyu2022mim,
  title={MIM: A deep mixed residual method for solving high-order partial differential equations},
  author={Lyu, Liyao and Zhang, Zhen and Chen, Minxin and Chen, Jingrun},
  journal={Journal of Computational Physics},
  volume={452},
  pages={110930},
  year={2022},
  publisher={Elsevier}
}

@article{li2024priori,
  title={A priori error estimate of deep mixed residual method for elliptic pdes},
  author={Li, Lingfeng and Tai, Xue-Cheng and Yang, Jiang and Zhu, Quanhui},
  journal={Journal of Scientific Computing},
  volume={98},
  number={2},
  pages={44},
  year={2024},
  publisher={Springer}
}

@misc{driscoll2014chebfun,
  title={Chebfun guide},
  author={Driscoll, Tobin A and Hale, Nicholas and Trefethen, Lloyd N},
  year={2014},
  publisher={Pafnuty Publications, Oxford}
}

@article{lagaris1998artificial,
  title={Artificial neural networks for solving ordinary and partial differential equations},
  author={Lagaris, Isaac E and Likas, Aristidis and Fotiadis, Dimitrios I},
  journal={IEEE transactions on neural networks},
  volume={9},
  number={5},
  pages={987--1000},
  year={1998},
  publisher={IEEE}
}

@article{igelnik2003kolmogorov,
  title={Kolmogorov's spline network},
  author={Igelnik, Boris and Parikh, Neel},
  journal={IEEE transactions on neural networks},
  volume={14},
  number={4},
  pages={725--733},
  year={2003},
  publisher={IEEE}
}

@book{kolmogorov1961representation,
  title={On the representation of continuous functions of several variables by superpositions of continuous functions of a smaller number of variables},
  author={Kolmogorov, Andre{\u\i} Nikolaevich},
  year={1961},
  publisher={American Mathematical Society}
}

@article{arnol1959representation,
  title={On the representation of continuous functions of three variables by superpositions of continuous functions of two variables},
  author={Arnol'd, Vladimir Igorevich},
  journal={Matematicheskii Sbornik},
  volume={90},
  number={1},
  pages={3--74},
  year={1959},
  publisher={Russian Academy of Sciences, Steklov Mathematical Institute of Russian~…}
}
\bibliographystyle{iclr2025_conference}

\appendix
\section{Explicit expression of functions}\label{Appendix: details of approximate functions}
The following functions are used in \cref{table: function approxmation}.
\begin{enumerate}
    \item \textit{sin-low}
    \begin{equation}
        f(x) = \sin (4\pi x),\quad x\in [-1,1]
        \end{equation}
    \item \textit{sin-high}
        \begin{equation}
        f(x) = \sin (400\pi x),\quad x\in [-1,1]
        \end{equation}
    \item \textit{bl}
        \begin{equation}
        f(x) = e^{-100x},\quad x\in [0,1]
        \end{equation}
    \item \textit{sqrt}
        \begin{equation}
        f(x) = \sqrt{x},\quad x\in [0,1]
        \end{equation}
    \item \textit{double-exponential}
    \begin{equation}
        f(x)=\frac{x(1-x) \mathrm{e}^{-x}}{(1 / 2)^2+(x-1 / 2)^2}, \quad x \in[0,1]
    \end{equation}
    \item \textit{multi-sqrt}
        \begin{equation}
        f(x) = x^{1/2}(1-x)^{3/4},\quad x\in [0,1]
        \end{equation}
    \item \textit{piece-wise}
        \begin{equation}
        f(x)=\left\{
        \begin{aligned}
         & \sin(20 \pi x) + x^2,&\quad x\in [0,0.5] \\
         & 0.5x e^{-x} + |\sin(5\pi x)|,&\quad x\in [0.5,1.5] \\
         & \log (x - 1) / \log(2) - \cos(2  \pi x),&\quad x\in [1.5,2]
        \end{aligned}
        \right.
        \end{equation}
    \item \textit{spectral-bias}
        \begin{equation}
        f(x)=\left\{
        \begin{aligned}
         & \sum_{k=1}^4\sin(k x) + 5,&\quad x\in [-1, 0] \\
         & \cos(10x),&\quad x\in [0, 1] \\
        \end{aligned}
        \right.
        \end{equation}
\end{enumerate}
\section{Details of PDEs}\label{Appendix: details of pde}
\subsection{1D problems}
\subsection{Perturbed boundary value problem}
We consider the singularly perturbed second-order boundary value problem (\textit{perturbed} in \cref{table:pikan_pde}):
\begin{equation}
    \epsilon u_{x x}-u_x=f(x), \quad x \in[-1,1].
\end{equation}
In specific cases, the problem has exact solutions, in this paper, we choose $f(x)=-1$, and the exact solution is
\begin{equation}
    u(x)=1+x+\frac{e^{\frac{x}{\epsilon}}-1}{e^{\frac{1}{\epsilon}}-1},
\end{equation}
where $\epsilon=0.01$ in our experiments.
\subsection{Nonlinear problem}
We consider the nonlinear boundary value problem (\textit{nonlinear} in \cref{table:pikan_pde}):
\begin{equation}
\begin{aligned}
        -u_{xx}+\frac{u_x}{x}+\frac{u}{x^2} & =\frac{\left(-41 x^2+34 x-1\right)\sqrt{x}}{4}-2 x+\frac{1}{x^2}, \quad x\in[0,1] \\
        u(0)-2u_x(0) & =1,\\
        3u(1)+u_x(1) & =9,
\end{aligned}
\end{equation}
with the exact solutions
\begin{equation}
    u(x)=x^{5 / 2}(1-x)^2+x^3+1.
\end{equation}
\subsection{Burgers}
We consider the Burgers' equation (\textbf{Burgers' equation} in \cref{table:ablation}):
\begin{equation}\label{eq:burgers}
\frac{\partial u}{\partial t}+u \frac{\partial u}{\partial x}-\nu \frac{\partial^2 u}{\partial x^2}=0, \quad x\in [-1,1], t\in [0,0.1].
\end{equation}
with Dirichlet boundary condition, and the exact solution is
\begin{equation}
    u=\frac{a}{2} - \frac{a\tanh\left(\frac{a (x - a  t / 2) }{4\nu}\right)}{2},
\end{equation}
where $a=0.5, \nu=0.01$ in our experiments. 
\subsection{T-nonlinear problem}\label{eq:tnonlinear}
We consider the time-dependent nonlinear problem (\textbf{T-nonlinear} in \cref{table:ablation}):
\begin{equation}
\begin{gathered}
     u_t=\frac{x+2}{t+1}u_x, \quad x\in [-1,1], t \in [0,0.1].\\
     u(x,0)=\cos(x+2),\\
     u(1,t)=\cos(3(t+1)),
\end{gathered}
\end{equation}
with the exact solution:
\begin{equation}
    u(x,t)=\cos((t+1)(x+2)).
\end{equation}
\subsection{Convection-diffusion}
We consider the 1-D convection-diffusion equation with periodic boundary conditions (used in \cref{Appendix: oscillation}):
\begin{equation}\label{eq:cd}
\begin{gathered}
    u_t+au_x-\epsilon u_{xx}=0, x\in \left[-1,1\right], t \in \left[0,0.1\right],\\
    u(x,0)=\sum_{k=0}^{5} \sin\left(k \pi x\right),
\end{gathered}
\end{equation}
with the analytic solution 
\begin{equation}
    u(x,t)=\sum_{k=0}^{5} \sin\left(k\pi x - ka\pi t\right)  e^{-\epsilon  k^2 \pi^2 t},
\end{equation}
  where $\epsilon=0.01$, and $a=0.1$in our experiments.

\subsection{2D problems}
\subsubsection{Boundary layer}
We consider the 2-D boundary layer problem (\textit{bl-2d} in \cref{table:pikan_pde}):
\begin{equation}\label{eq:bl2d}
u_{xx}/\alpha_1+u_{x}+u_{yy}/\alpha_2+u_{y}=0 ,
\end{equation}
with the exact solution
\begin{equation}
   u(x,y)=\exp (-\alpha_1 x)+\exp (-\alpha_2 y),
\end{equation}
where $\alpha_1=\alpha_2=100$ in our experiments. 

\begin{figure}[ht]
  \centering 
  \subfigure[\label{bl_2d_test}]{\includegraphics[width=0.32\linewidth]{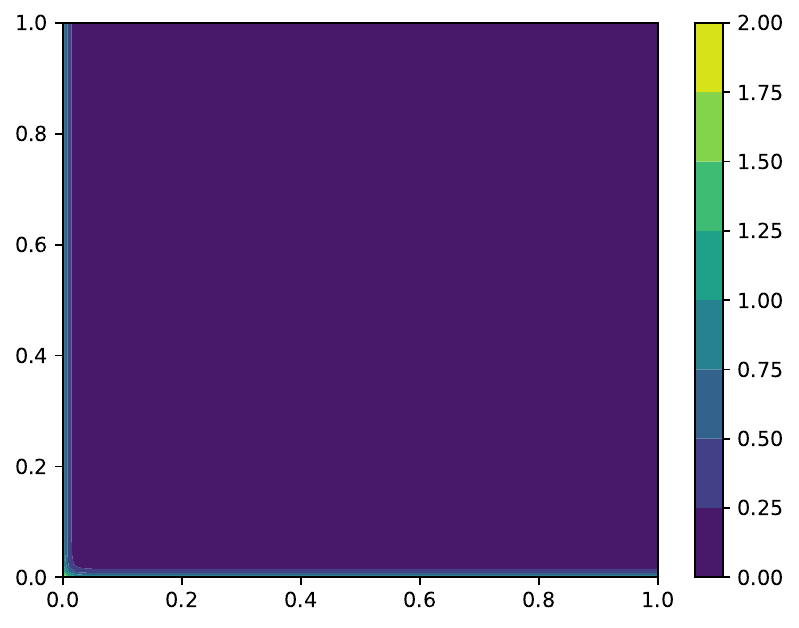}}
    \subfigure[\label{bl_2d_pred}]{\includegraphics[width=0.32\linewidth]{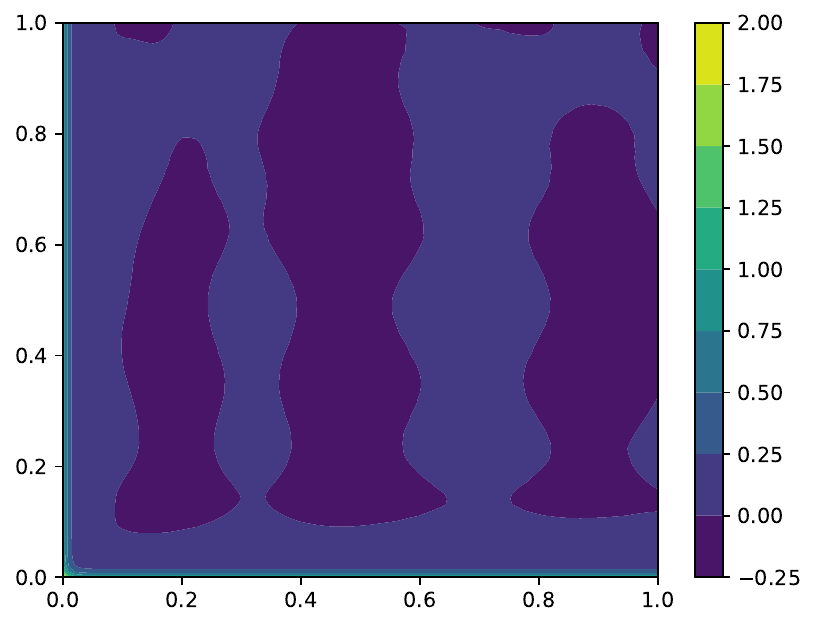}}
    \subfigure[\label{bl_2d_error}]{\includegraphics[width=0.32\linewidth]{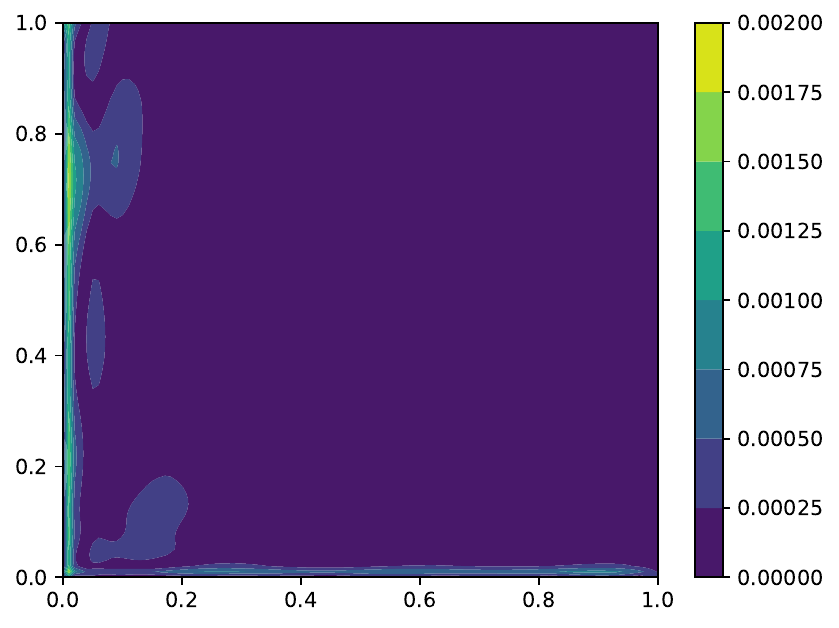}}
  \caption{\cref{bl_2d_test} depicts the exact solution of \cref{eq:bl2d}, \cref{bl_2d_pred} shows the solution predicted by SincKAN, \cref{bl_2d_error} shows the absolute error between the predicted solution and the exact solution, exhibits that the error mainly comes from the boundary layer.}
  \label{fig:bl_2d}
\end{figure}
\subsubsection{Navier stokes equations}
We consider the Taylor–Green vortex (\textit{ns-tg-u} and \textit{ns-tg-v} in \cref{table:pikan_pde}):
\begin{equation}
    \begin{gathered}
        \nabla \cdot \boldsymbol{u}=0, \quad t\in [0,T], \ \boldsymbol{x} \in \Omega,\\
        \partial_t \boldsymbol{u}+\boldsymbol{u} \cdot \nabla \boldsymbol{u}=-\nabla p+\nu\triangle \boldsymbol{u},\quad t\in [0,T], \ \boldsymbol{x} \in \Omega,\\
    \end{gathered}
\end{equation}
where $\boldsymbol{u}=(u,v)$,with the exact solution
\begin{equation}
    \begin{gathered}
            u = -\cos(x) \sin(y) \exp(-2 \nu t) \\
    v = \sin(x) \cos(y) \exp(-2 \nu t) \\
    p = -\left(\cos(2 x) + \sin(2 y)\right)  \exp(-4  \nu t) / 4
    \end{gathered}
\end{equation}
with $T=1$, $\nu=1/400$ in our experiments. After dimensionless, $\boldsymbol{x} \in [0,1]^2$.
\section{Experiment details}\label{Appendix: expeirment details}
Totally, in our experiments, the Adam~\cite{kingma2014adam} optimizer is used with the exponential decay learning rate. The MLP and modified MLP are equipped with the tanh activations and Xavier initialization inherited from \cite{raissi2019physics}. 
\subsection{Approximation}
The hyperparameters of used networks are shown in \cref{table:cost for kans}.
\begin{itemize}
    \item For the 1-D problem, we generate the training dataset by uniformly discretizing the input interval to 5000 points and train the network with 3000 points randomly sampled from the training dataset for each iteration. In total, We train every network with $10^5$ iterations. Additionally, to evaluate the generalization, we generate the testing (fine) dataset by uniformly discretizing the input interval to 10000 points.
\end{itemize}
\subsection{PIKANs}
The hyperparameters of used networks are shown in \cref{table:cost for pikans}. 
\begin{itemize}
    \item For time-independent 1-D problems, we generate the training dataset by uniformly discretizing the input interval to 1000 points, then train the network with 500 points randomly sampled from the training dataset for each iteration. In total, We train every network with $1.5\times 10^6$ iterations.
    \item For time-dependent 1-D problems, we generate the training dataset by uniformly discretizing the spatial dimension to 1000 points and the temporal dimension to 11 points, then train the network with 5000 points randomly sampled from the training dataset for each iteration. In total, We train every network with $1.5\times 10^6$ iterations.
    \item  For time-independent 2-D problems, we generate the training dataset by uniformly discretizing every dimension to 100 points, then train the network with 5000 points randomly sampled from the training dataset for each iteration. In total, We train every network with $1.5\times 10^6$ iterations.
    \item  For time-dependent 2-D problems, we generate the training dataset by uniformly discretizing every spatial dimension to 100 points and the temporal dimension to 11 points, then train the network with 50000 points randomly sampled from the training dataset for each iteration. In total, We train every network with $1.5\times 10^6$ iterations.
\end{itemize}
\section{Interior approximation of SincKAN}\label{Appendix: interior}
\begin{figure}[ht]
   \centering
    \subfigure[\label{interior_approx}\cref{eq:bl2d} for approximation]{\includegraphics[width=0.6\linewidth]{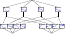}}
    \subfigure[\label{interior_pde}\cref{eq:bl2d} for PIKANs]{\includegraphics[width=0.6\linewidth]{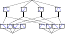}}
   
   \caption{\cref{interior_approx} used high $M$ so the interpolated $phi$ has oscillations while \cref{interior_pde} uses low $M$ so the $phi$ is more smooth}
    \label{fig:interior}
\end{figure}
\section{Approximation on fine grids} \label{Appendix: fine grids}
As we discussed in \cref{Appendix: expeirment details}, we additionally evaluate every network on fine grids. And due to the oscillations discussed in \cref{Appendix: oscillation}, \cref{table: function fine grids} reveals the weak generalization of SincKANs demands further research, although the applications of approximating a function don't strongly require this capability.
\begin{table}[ht]
\caption{RMSE evaluated on fine grids}
\label{table: function fine grids}
\begin{center}
\begin{adjustbox}{width=\columnwidth, center}
\begin{tabular}{llllll}
\multicolumn{1}{c}{\bf Function name}  & \multicolumn{1}{c}{\bf MLP} & \multicolumn{1}{c}{\bf modified MLP} &\multicolumn{1}{c}{\bf KAN} & \multicolumn{1}{c}{\bf ChebyKAN} & \multicolumn{1}{c}{\bf SincKAN (ours)} 
\\ \hline \\
\textit{sin-low} & $1.51 e\mbox{-}2 \pm 2.01 e\mbox{-}2$ & $7.29e\mbox{-}4 \pm 2.97 e\mbox{-}4$ & $1.27e\mbox{-}3 \pm 3.04e\mbox{-}4$ &
$1.76e\mbox{-}3 \pm 3.19e\mbox{-}4$ &
\textcolor{red}{$4.46 e\mbox{-}4 \pm 2.79 e\mbox{-}4$} \\ 
\textit{sin-high} & $7.07e\mbox{-}1 \pm 4.21e\mbox{-}8$ & $7.07 e\mbox{-}1 \pm 1.31 e\mbox{-}5$ & $7.06e\mbox{-}1 \pm 1.29e\mbox{-}3$ &
$5.70e\mbox{-}2 \pm 5.99 e\mbox{-}3$ &
\textcolor{red}{$4.15e\mbox{-}2 \pm 4.53e\mbox{-}3$} \\ 
\textit{bl} & $7.63e\mbox{-}4 \pm 1.13e\mbox{-}3$ & $5.72e\mbox{-}4 \pm 4.01e\mbox{-}4$ & $2.62e\mbox{-}4 \pm 7.89e\mbox{-}5$ &
$1.81e\mbox{-}3 \pm 6.98e\mbox{-}4$ &
\textcolor{red}{$2.28e\mbox{-}4 \pm 1.16e\mbox{-}4$} \\ 
\textit{double exponential} & $1.95e\mbox{-}3 \pm 8.17e\mbox{-}4$ & $7.76e\mbox{-}5 \pm 4.07e\mbox{-}5$ & $2.18e\mbox{-}4 \pm 1.51e\mbox{-}4$ &
$3.11e\mbox{-}3 \pm 2.16 e\mbox{-}3$ &
\textcolor{red}{$7.06e\mbox{-}5 \pm 1.09e\mbox{-}5$} \\ 
sqrt & $3.06e\mbox{-}3 \pm 9.34 e\mbox{-}4$ & \textcolor{red}{$4.46 e\mbox{-}5 \pm 5.51e\mbox{-}5$} & $4.79e\mbox{-}4 \pm 1.23e\mbox{-}4$ &
$3.69e\mbox{-}3 \pm 1.27 e\mbox{-}3$ &
$3.24e\mbox{-}4 \pm 1.31e\mbox{-}4$\\ 
\textit{multi-sqrt} & $2.06e\mbox{-}3 \pm 1.16e\mbox{-}3$ & $4.59e\mbox{-}4 \pm 4.86e\mbox{-}4$ & $3.61e\mbox{-}4 \pm 8.67 e\mbox{-}5$ &
$2.34e\mbox{-}3 \pm 1.17e\mbox{-}3$ &
\textcolor{red}{$2.14e\mbox{-}4 \pm 2.49 e\mbox{-}4$} \\ 
\textit{piece-wise} & $2.06e\mbox{-}2 \pm 5.57e\mbox{-}3$ & $3.75e\mbox{-}2 \pm 1.81e\mbox{-}2$ & $5.84e\mbox{-}2 \pm 1.03e\mbox{-}2$ &
\textcolor{red}{$7.28e\mbox{-}3 \pm 9.59 e\mbox{-}4$} &
$9.41e\mbox{-}3 \pm 2.14e\mbox{-}4$ \\ 
\textit{spectral-bias} & $2.48e\mbox{-}2 \pm 9.77 e\mbox{-}3$ & \textcolor{red}{$1.88e\mbox{-}2 \pm 9.55e\mbox{-}4$} & $4.79e\mbox{-}2 \pm 9.44 e\mbox{-}3$ &
$2.18e\mbox{-}2 \pm 2.97 e\mbox{-}4$ &
$2.21e\mbox{-}2 \pm 9.98 e\mbox{-}5$ \\ 
\end{tabular}
\end{adjustbox}
\end{center}
\end{table}
\section{Details of other networks}\label{Appendix: networks}
\subsection{MLP}
Multilayer Perceptron (MLP) is the neural network consisting of fully connected neurons with a nonlinear activation function, and can be represented simply by:
\begin{equation}
\operatorname{MLP}(x)=\left(W^{L-1} \circ \sigma \circ W^{L-2} \circ \sigma \circ \cdots \circ W^1 \circ \sigma \circ W^0\right) \vx
\end{equation}
where $W^i(x)=W_ix+b_i$, $W_i \in \mathbb{R}^{m_i \times n_i}$ is a learnable matrix, $b_i \in \mathbb{R}^{m_i}$ is a learnable bias, $\sigma$ is the chosen nonlinear activation function, and $L$ is the depth of MLP.
\subsection{Modified MLP}
Modified MLP is an upgraded network of MLP inspired by the transformer networks. It introduces two extra features and has a skip connection with them:
\begin{equation}
\begin{gathered}
    U=\sigma(W^{L+1}\vx), \quad, V=\sigma(W^{L+2}\vx), \quad H^1=\sigma(W^{0}\vx), \\
    H^{i+1} = (1-\sigma(W^{i}H^{i}))*U+\sigma(W^{i}H^{i})*V, \quad i=1,\cdots,L, \\
    \operatorname{Modified MLP}(\vx) = W^{L+3}H^{L+1},
\end{gathered}
\end{equation}
where $*$ is the element-wise multiplication.
\subsection{KAN}
 Kolmogorov-Arnold Network (KAN) is a novel network proposed recently that aims to be more accurate and interpretable than
MLP. The main difference is KAN's activation functions are learnable: suppose $\mathbf{\Phi}=\{\phi_{p,q}\}$ is the matrix of univariable functions where $p= 1, 2, \dots n_{in}, q= 1, 2, \dots n_{out}$ and $\theta$ represents the trainable parameters. The KAN can be defined by:
\begin{equation}
    \operatorname{KAN} (\vx) = \left(\mathbf{\Phi}_{L-1} \circ \mathbf{\Phi}_{L-2} \circ \cdots \circ \mathbf{\Phi}_1 \circ \mathbf{\Phi}_0\right) \vx, \quad \vx\in \mathbb{R}^d,
\end{equation}
where 
\begin{equation}\label{eq: def phi}
\mathbf{\Phi}_{l}(\vx^{(l)})= \left\{\sum_{i=1}^{n^{(l)}_{in}} \phi_{j, i}\left(x_{i}^{(l)}\right)\right\}_{j=1}^{n^{(l)}_{out}},  \quad \forall l = 0,1,\cdots,L-1
\end{equation}
where $\vx \in \mathbb{R}^{n_{in}^{(l)}}, \mathbf{\Phi}_{l}(\vx^{(l)}) \in \mathbb{R}^{n_{out}^{(l)}}$.
To approximate every single activation function $\phi$, KAN utilizes the summation of basis function and spline interpolation:
\begin{equation}
    \phi\left(x\right)=w_b \mathrm{silu}\left(x\right)+w_s \left(\sum_i c_i B_i(x)\right),
\end{equation}
where $c_i, w_s, w_b$ are learnable, and $B_i$ is the spline.
\subsection{ChebyKAN}
ChebyKAN utilizes the Chebyshev polynomials to construct the learnable activation function $\phi$ in KAN. And the modified ChebyKAN embeds the $\tanh$ activation function between every layer. Thus the ChebyKAN used in our experiments can be defined by:
\begin{equation}
    \operatorname{ChebyKAN} (\vx) = \left(\mathbf{\Phi}_{L-1} \circ \tanh \circ \mathbf{\Phi}_{L-2}\circ \cdots \circ \mathbf{\Phi}_1 \circ \tanh \circ \mathbf{\Phi}_0\circ \tanh \circ\right) \vx, \quad \vx\in \mathbb{R}^d,
\end{equation}
where $\mathbf{\Phi}$ has the same definition of \cref{eq: def phi} with different univariable function
\begin{equation}
    \phi = \sum_i c_i T_i(x),
\end{equation}
where $T_i$ is the $i$th Chebyshev polynomial.
\section{Computational Cost}\label{Appendix: cost}
\subsection{Training}
In \cref{eq: Sinc activation function}, SincKAN has an additional summation on several $h$, so the trainable coefficients $c$ are $M$ times larger than KAN and ChebyKAN. However, the training time is not only dependent on the number of total parameters, thus, we demonstrate the cost of training for approximation in \cref{table:cost for kans}, and demonstrate the cost of training for PDEs in \cref{table:cost for pikans}.  In \cref{table:cost for kans} and \cref{table:cost for pikans}, we use '$\text{depth}\times\text{width}$' to represent the size for MLP and modified MLP; '$\text{width} \times \text{degree} $' to represent the size for KAN and ChebyKAN; and '$\text{width} \times \text{degree} \times M $' to represent the size for SincKAN. Note that we train the network in two environments distinguished by two superscripts:

\dag: training on single NVIDIA A100-SXM4-80GB with CUDA version: 12.4.

\ddag: training on single NVIDIA A40-48GB with CUDA version: 12.4.

\begin{table}[ht]
\caption{Computational cost for approximation$^\dag$}
\label{table:cost for kans}
\begin{center}
\begin{tabular}{lllll}
\multicolumn{1}{c}{\bf Network} & \multicolumn{1}{c}{\bf Size}  & \multicolumn{1}{c}{\bf Training rate (iter/sec)} 
& \multicolumn{1}{c}{\bf Referencing time (ms)} 
& \multicolumn{1}{c}{\bf Parameters}\\ 
\hline \\
MLP & $10\times 100$ & $9.89\times 10^{2}$ & 
$1.62\times 10^{1}$ & 
$81101$\\ 
modified MLP & $10\times 100$ & $9.13\times 10^{2}$ &
$2.92\times 10^{1}$ &
$81501$\\ 
KAN & $8\times 8$ & $1.15\times 10^{3}$ &
$1.01\times 10^{2}$ &
$160$\\ 
ChebyKAN & $40\times 40$ & $1.29\times 10^{3}$ &
$3.11\times 10^{1}$ &
$3280$  \\ 
SincKAN & $8\times 100 \times 6$ & $1.29\times 10^{3}$ & 
$2.06\times 10^{1}$ &
$9696$  \\ 
\end{tabular}
\end{center}
\end{table}
\begin{table}[ht]
\caption{Computational cost for train PIKANs}
\label{table:cost for pikans}
\begin{center}
\begin{tabular}{lllll}
\multicolumn{1}{c}{\bf Function name}  & \multicolumn{1}{c}{\bf Network}& \multicolumn{1}{c}{\bf Size}  & \multicolumn{1}{c}{\bf Training rate (iter/sec)} & \multicolumn{1}{c}{\bf Parameters} \\ 
\hline \\
\multirow{5}{*}{boundary layer$^\ddag$ } & MLP &$10\times 100$ & $6.47\times 10^{2}$ & 81101 \\ 
                      & modified MLP & $10\times 100$ &$3.55\times 10^{2}$ & 81501\\ 
                      & KAN & $8\times 8$  &$1.33\times 10^{3}$ & 160\\ 
                      & ChebyKAN & $40\times 40$ & $1.89\times 10^{3}$  & 3280\\ 
                      & SindcKAN & $8\times 8 \times 1$ & $1.27\times 10^{3}$  & 194 \\ 
\\
\multirow{5}{*}{\textit{perturbed}$^\ddag$ } & MLP & $10\times 100$ & $6.85\times 10^{2}$ & 81101 \\ 
                      & modified MLP &$10\times 100$ & $3.59\times 10^{2}$ & 81501\\ 
                      & KAN & $8\times 8$ & $1.27\times 10^{3}$ & 160\\ 
                      & ChebyKAN & $40\times 40$ & $1.07\times 10^{3}$ & 3280 \\ 
                      & SincKAN & $8\times 8 \times 1$ & $1.25\times 10^{3}$ &194 \\ 
\\
\multirow{5}{*}{\textit{nonlinear}$^\ddag$ } & MLP & $10\times 100$ & $4.62\times 10^{2}$ & 81101\\ 
                      & modified MLP & $10\times 100$ & $3.07\times 10^{2}$ & 81501\\ 
                      & KAN & $8\times 8$ & $1.56\times 10^{3}$ & 160\\ 
                      & ChebyKAN & $40\times 40$ & $1.53\times 10^{3}$ & 3280 \\ 
                      & SincKAN & $8\times 4 \times 1$ & $1.54\times 10^{3}$ & 130 \\ 
\\
\multirow{5}{*}{\textit{bl-2d}$^\ddag$ } & MLP & $10\times 100$ & $2.39\times 10^{2}$ & 81201\\ 
                      & modified MLP & $10\times 100$ & $1.96\times 10^{2}$ & 81801\\ 
                      & KAN & $8\times 8$ & $3.46\times 10^{2}$ & 240\\ 
                      & ChebyKAN & $40\times 40$ & $4.95\times 10^{2}$ & 4920 \\ 
                      & SincKAN & $8\times 20 \times 1$ & $2.97\times 10^{2}$ & 570 \\ 
\\
\multirow{5}{*}{\textit{ns-tg}$^\dag$ } & MLP & $10\times 100$ & $1.71\times 10^{2}$ & 81503\\ 
                      & modified MLP & $10\times 100$ & $1.51\times 10^{2}$ & 82303\\ 
                      & KAN & $8\times 8$ & $2.55\times 10^{2}$ & 480\\ 
                      & ChebyKAN & $40\times 40$ & $3.83\times 10^{3}$ & 9840 \\ 
                      & SincKAN & $8\times 8 \times 1$ & $2.77\times 10^{2}$ & 550 \\ 
\end{tabular}
\end{center}
\end{table}

\subsection{Referencing}
As the referencing cost doesn't depend on the task \textit{i.e.} the loss function, the results are evaluated on the model trained by approximation task and the results can be found in \cref{table:cost for kans}. The results reveal that although SincKAN has much more parameters than KAN and ChebyKAN, SincKAN is faster when referencing. Note that the referencing is slower than training because we compile the training procedure by JAX~\cite{jax2018github}.
\section{Results of different degree}\label{Appendix: scaling law}
\begin{figure}[ht]
  \centering 
  \subfigure[\label{degree_100} $N_{points}=100$]{\includegraphics[width=0.3\linewidth]{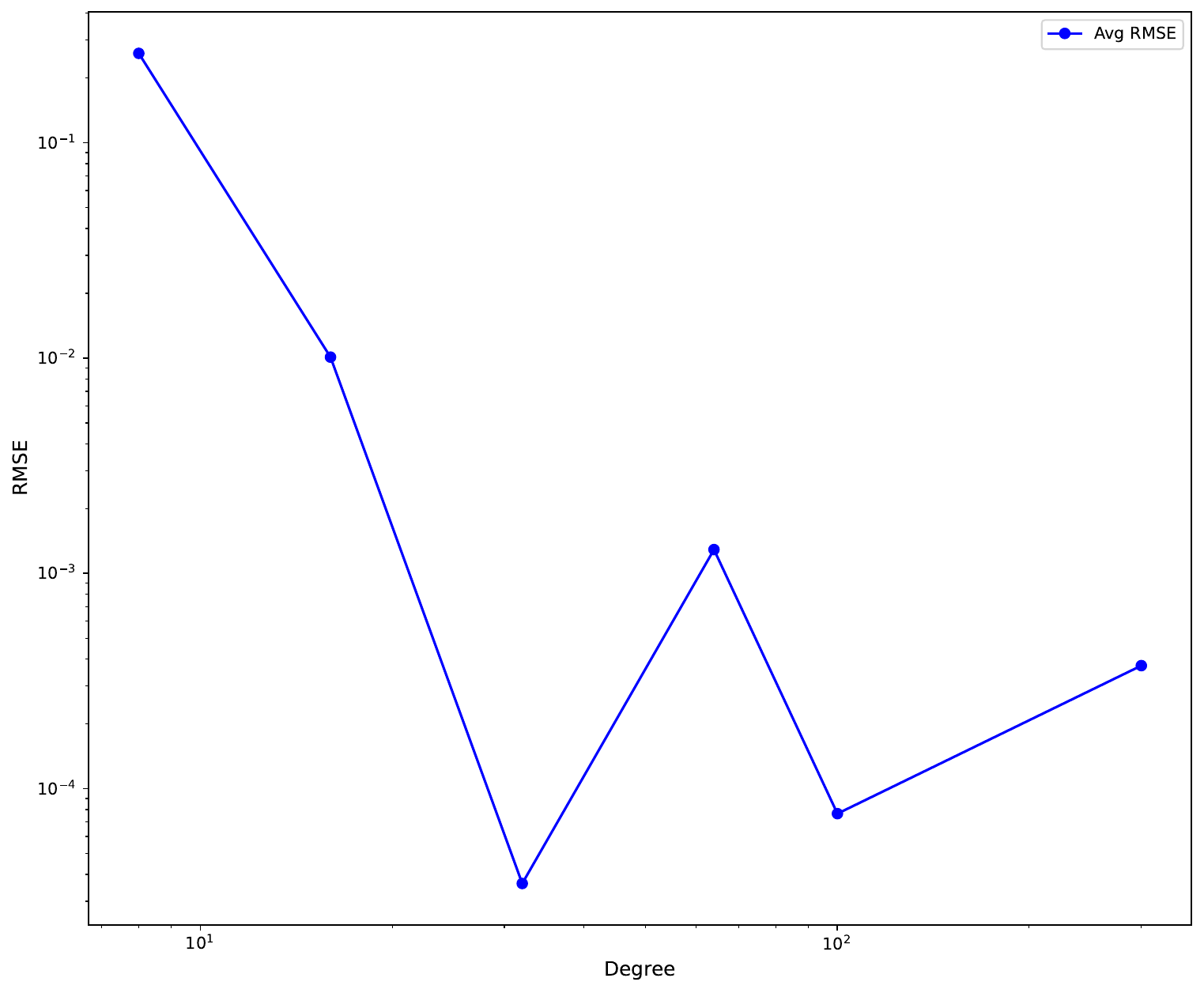}} 
  \subfigure[\label{degree_500} $N_{points}=500$]{\includegraphics[width=0.3\linewidth]{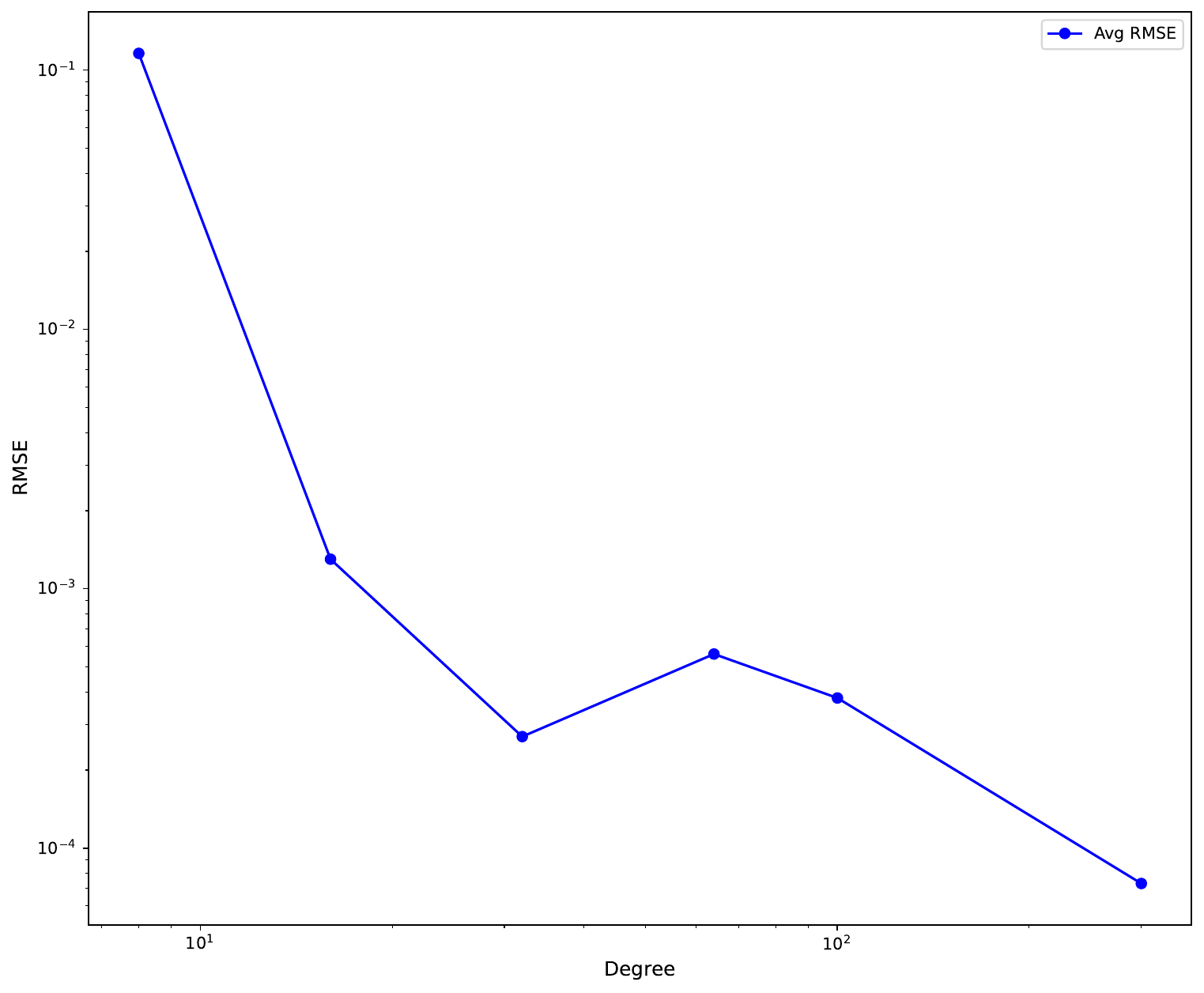}}
  \subfigure[\label{degree_1000} $N_{points}=1000$]{\includegraphics[width=0.3\linewidth]{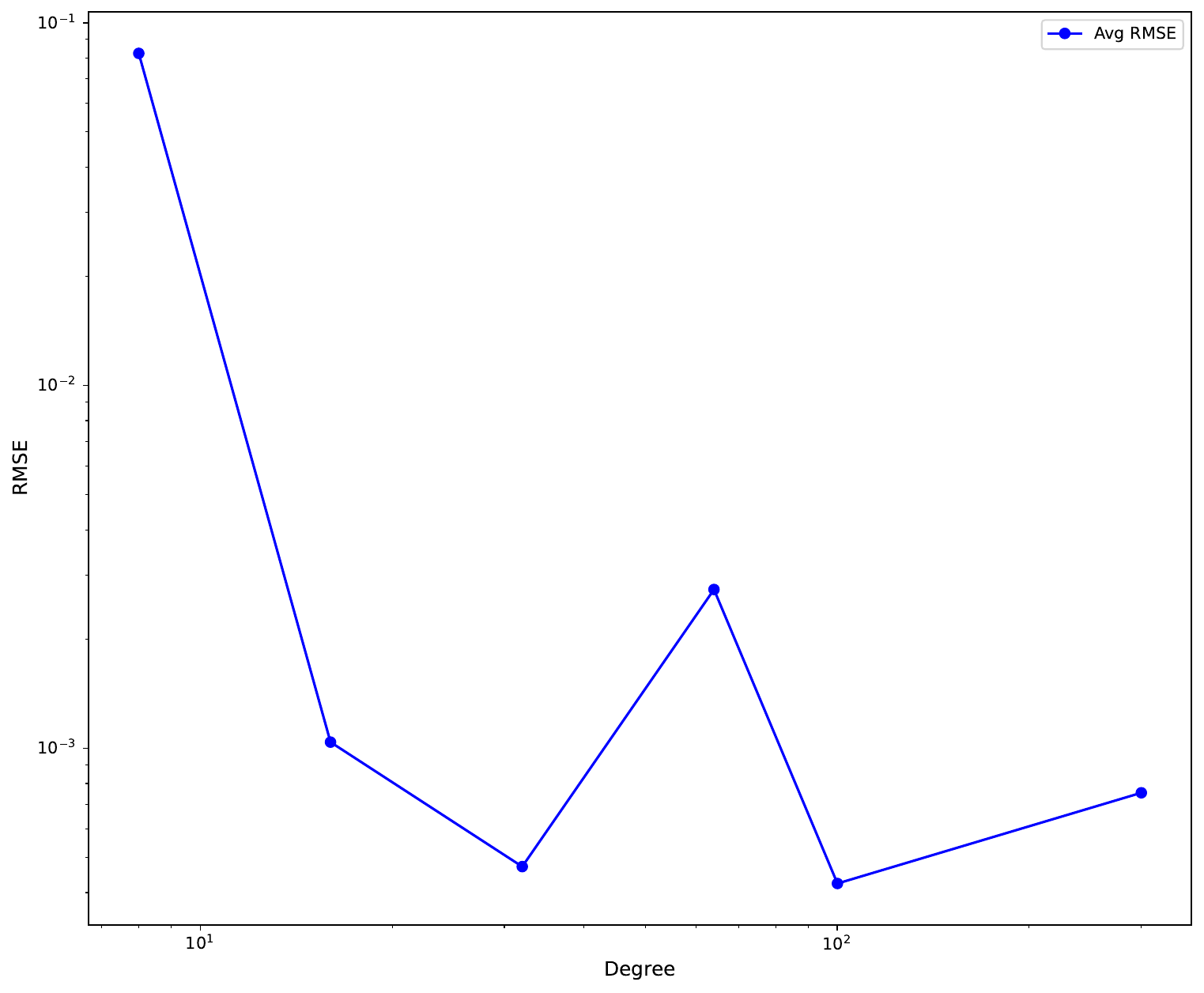}}
    \subfigure[\label{degree_5000} $N_{points}=5000$]{\includegraphics[width=0.3\linewidth]{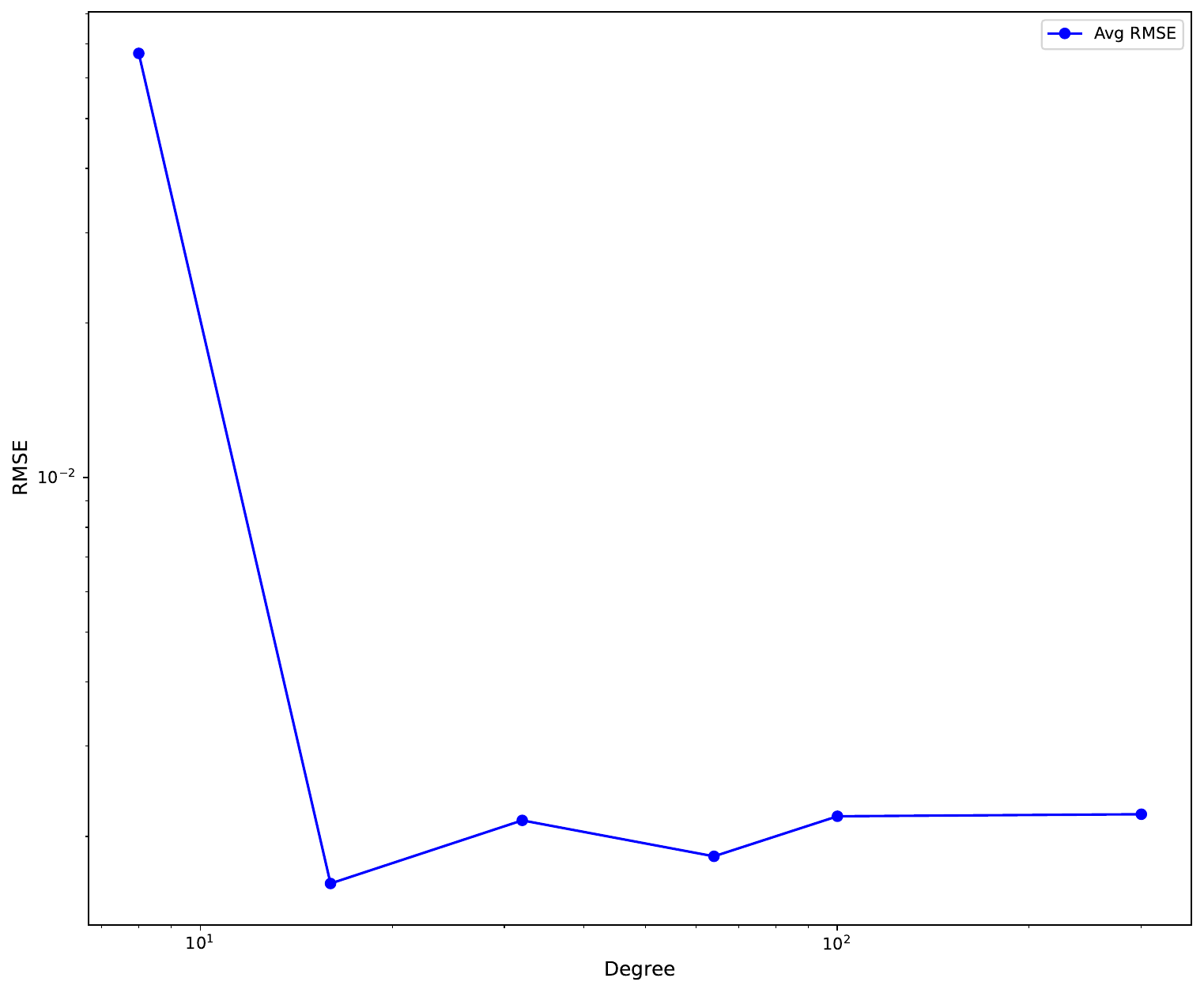}}
        \subfigure[\label{degree_10000} $N_{points}=10000$]{\includegraphics[width=0.3\linewidth]{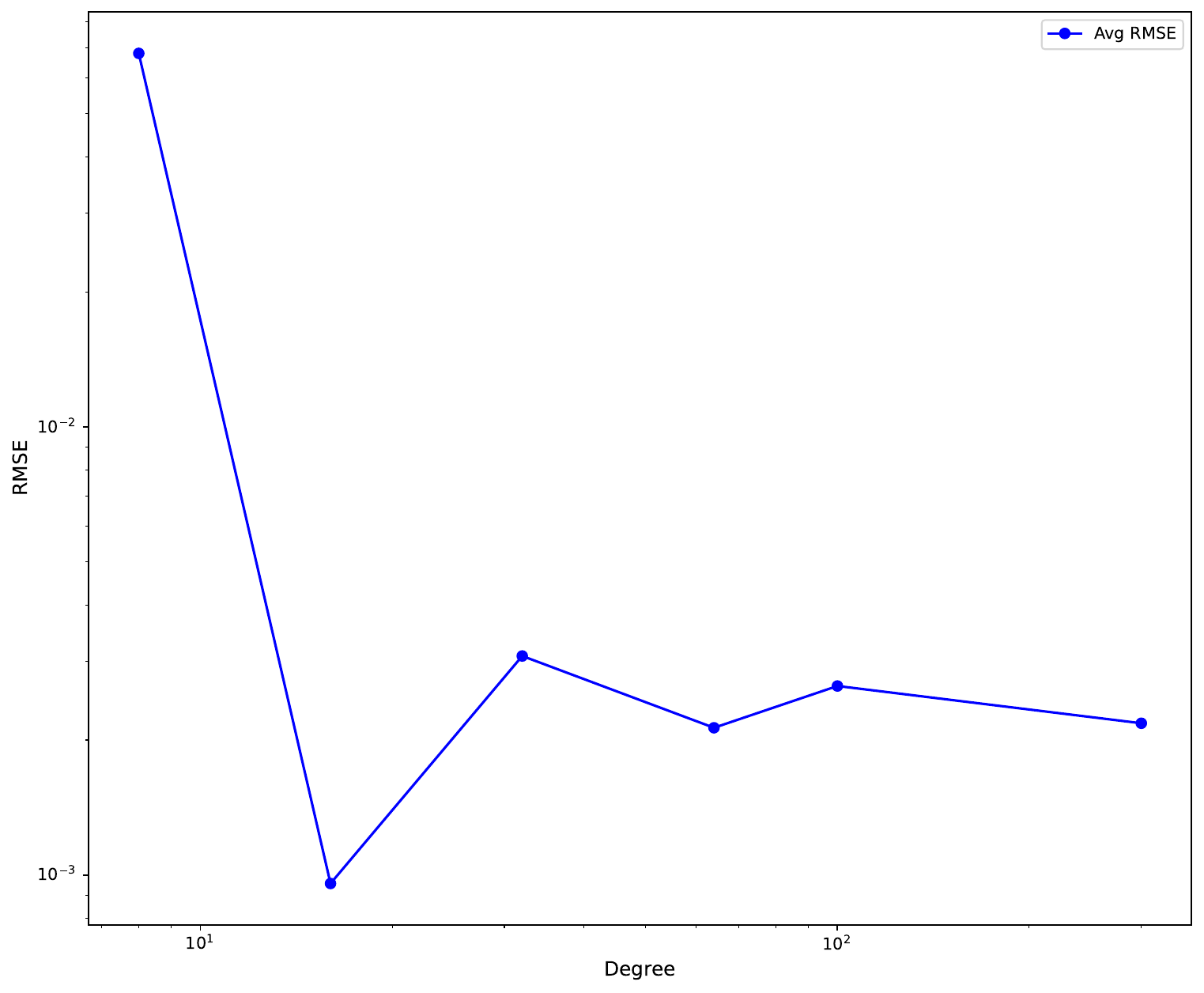}}
  \caption{Figures of different $N_{ponits}$ with increasing degree.}
  \label{fig:degree_points}
\end{figure}

In this experiment, we train our SincKAN on \textit{spectral-bias} function on $N=8,16,32,64,100,300$ and $N_{points}=100,500,1000,5000,10000$ with the inverse decay $\{h_i\}_{i=1}^M$ in $M=6$ and $h_0=7.0$.
Moreover, we set the batch size $N_{batch}=N_{points}/4$ to adapt to the changing of $N_{points}$. The results are shown in \cref{table: degree_points} and \cref{fig:degree_points}. Additionally, \cref{degree_100} shows that our neural scaling law is $\mathrm{RMSE} \propto G^{-4}$ compared to the best scaling law $\mathrm{RMSE} \propto G^{-3}$ claimed in KAN~\cite{liu2024kan}.
\begin{table}[ht]
\caption{RMSE for different degree and $N_{points}$}
\label{table: degree_points}
\begin{center}
\begin{adjustbox}{width=\columnwidth, center}
\begin{tabular}{clllll}
degree $\setminus$ $N_{points}$& \multicolumn{1}{c}{100} &\multicolumn{1}{c}{500} & \multicolumn{1}{c}{1,000} & \multicolumn{1}{c}{5,000} & \multicolumn{1}{c}{10,000}
\\ \hline \\
8 &$2.60e\mbox{-}1 \pm 2.76e\mbox{-}5$     & $1.16e\mbox{-}1 \pm 6.34e\mbox{-}6$   & $8.23e\mbox{-}2 \pm 4.54e\mbox{-}6$ & $6.70e\mbox{-}2 \pm 3.91e\mbox{-}3$ & $6.81e\mbox{-}2 \pm 4.51e\mbox{-}3$ \\
16                         & $1.01e\mbox{-}2 \pm 1.47e\mbox{-}2$  & $1.30e\mbox{-}3 \pm 9.42e\mbox{-}4$   & $1.04e\mbox{-}3 \pm 3.74e\mbox{-}4$ & $1.62e\mbox{-}3 \pm 2.48e\mbox{-}4$ & $9.57e\mbox{-}4 \pm 4.35e\mbox{-}4$ \\
32                         & \textcolor{red}{$3.63e\mbox{-}5 \pm 5.79e\mbox{-}5$}   & $2.69e\mbox{-}4 \pm 2.27e\mbox{-}4$   & $4.72e\mbox{-}4 \pm 2.75e\mbox{-}4$ & $2.15e\mbox{-}3 \pm 1.57e\mbox{-}3$ & $3.08e\mbox{-}3 \pm 7.14e\mbox{-}4$ \\
  64                       & $1.29e\mbox{-}3 \pm 2.21e\mbox{-}3$  & $5.60e\mbox{-}4 \pm 6.42e\mbox{-}4$   & $2.74e\mbox{-}3 \pm 4.01e\mbox{-}3$ & $1.83e\mbox{-}3 \pm 8.49e\mbox{-}4$ & $2.13e\mbox{-}3 \pm 9.05e\mbox{-}4$ \\
100 &$7.66e\mbox{-}5 \pm 1.16e\mbox{-}4$    & $3.79e\mbox{-}4 \pm 6.27e\mbox{-}4$  & $4.24e\mbox{-}4 \pm 7.42e\mbox{-}5$ & $2.19e\mbox{-}3 \pm 1.62e\mbox{-}3$ & $2.64e\mbox{-}3 \pm 1.76e\mbox{-}3$  \\ 
 300                        & $3.73e\mbox{-}4 \pm 5.62e\mbox{-}4$  &  $7.30e\mbox{-}5 \pm 4.25e\mbox{-}5$ & $7.54e\mbox{-}4 \pm 8.73e\mbox{-}4$ & $2.21e\mbox{-}3 \pm 1.04e\mbox{-}3$ & $2.18e\mbox{-}3 \pm 1.12e\mbox{-}3$  \\ 
\end{tabular}
\end{adjustbox}
\end{center}
\end{table}

\section{Results of selected h}\label{Appendix: select_h}
\cref{table: sin-low with inverse decay} and \cref{table: sin-low with exponential decay} show the results of selected $\{h_i\}$ in details. However, there are so many hyperparameters that may be adjusted when $h_i$ is larger. For example, for the large $h$ on fine grids, we argue that $N_{degree}=100$ may not exploit the capability fully. Thus, we conducted an extra experiment with 5000 grid points, $\{h_i\}=\{1/10,1/100,1/1000\}$, and $N_{degree}=500$. For sin-low, the RMSE is $7.92e\mbox{-}4\pm 4.21e\mbox{-}4 $, and for the sin-high, the RMSE is $2.32e\mbox{-}3\pm 2.74e\mbox{-}4 $. It shows that for sin-high, the SincKAN can obtain a more accurate result if we further tune the hyperparameters. 
\begin{figure}[ht]
  \centering 
  \subfigure[\label{sin_low_exp}]{\includegraphics[width=0.4\linewidth]{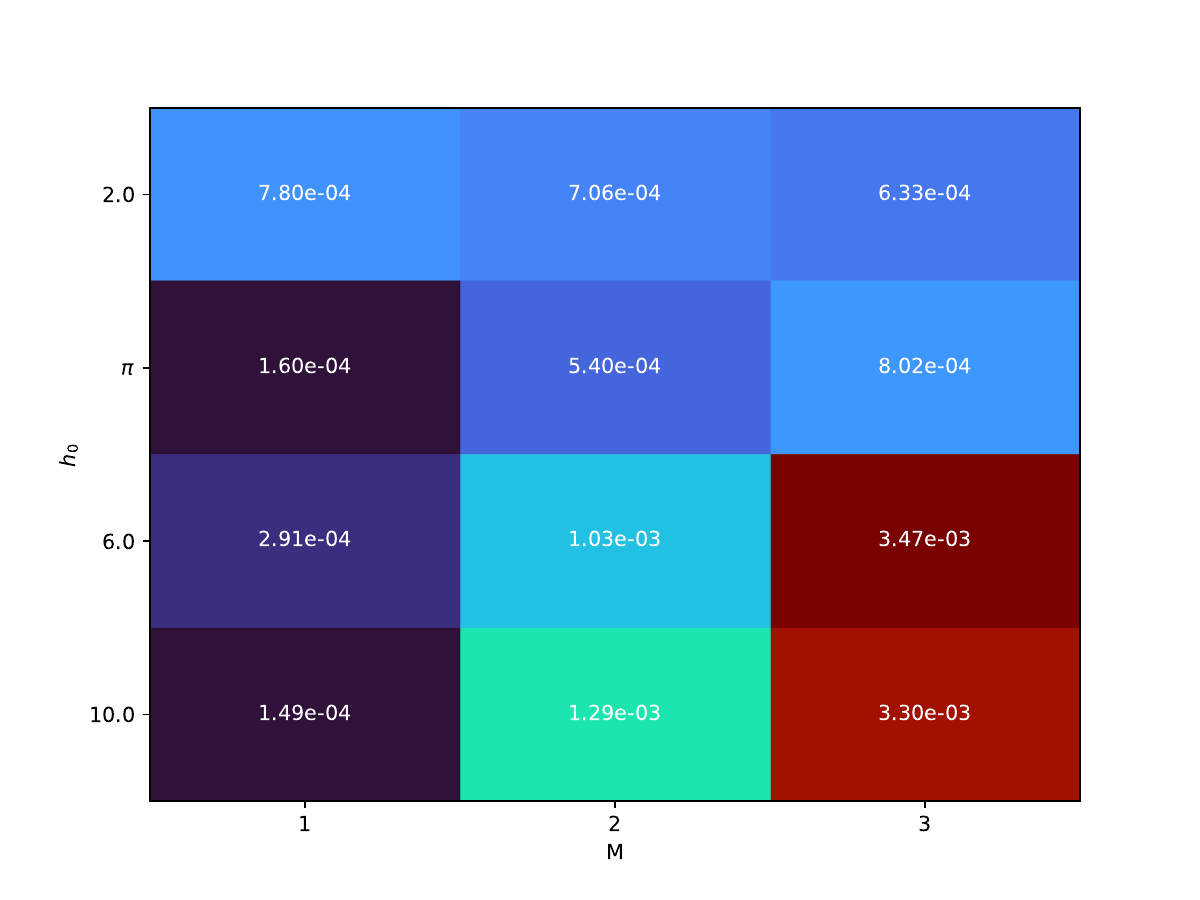}}
    \subfigure[\label{sin_high_exp}]{\includegraphics[width=0.4\linewidth]{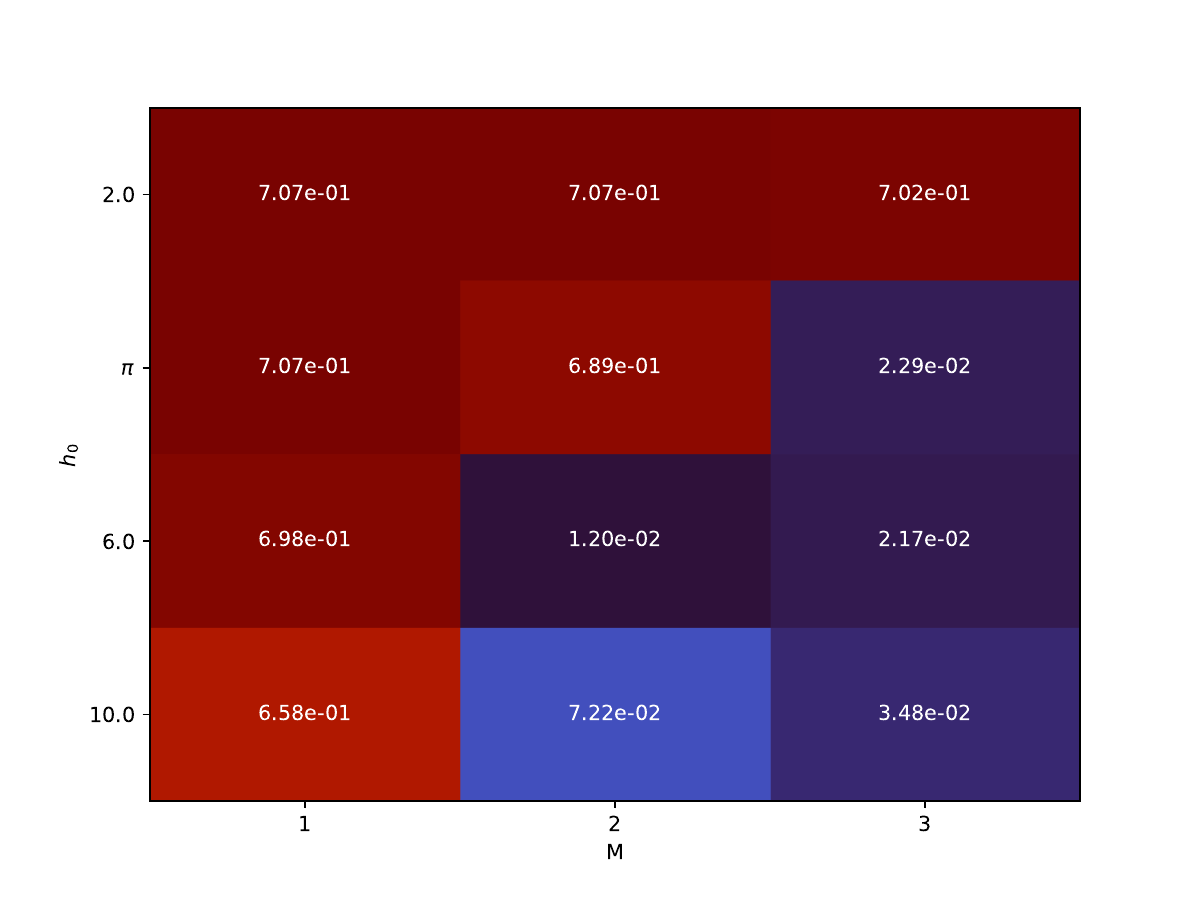}}
  \caption{\cref{sin_low_exp} shows the exponential decay approach on sin-low function; \cref{sin_high_exp} shows the exponential decay approach on sin-high function.}
  \label{fig:select_h_exp}
\end{figure}	
\begin{table}[ht]
\caption{RMSE for different $\{h_i\}$ with inverse decay}
\label{table: sin-low with inverse decay}
\begin{center}
\begin{adjustbox}{width=\columnwidth, center}
\begin{tabular}{llllll}
\multicolumn{1}{c}{\bf Function name}&$h_0$ $\setminus$ M& \multicolumn{1}{c}{1} &\multicolumn{1}{c}{6} & \multicolumn{1}{c}{12} & \multicolumn{1}{c}{24}
\\ \hline \\
\multirow{4}{*}{sin-low} &2.0     & $7.80e\mbox{-}4 \pm 7.96e\mbox{-}4$   & $2.43e\mbox{-}4 \pm 1.55e\mbox{-}4$ & $1.18e\mbox{-}3 \pm 2.38e\mbox{-}4$ & $4.30e\mbox{-}3 \pm 1.74e\mbox{-}3$ \\
                         & $\pi$  & $1.60e\mbox{-}4 \pm 6.66e\mbox{-}5$   & $8.96e\mbox{-}4 \pm 5.91e\mbox{-}4$ & $2.27e\mbox{-}3 \pm 8.53e\mbox{-}4$ & $3.81e\mbox{-}3 \pm 2.47e\mbox{-}3$ \\
                         & 6.0    & $2.91e\mbox{-}4 \pm 1.22e\mbox{-}4$   & $4.70e\mbox{-}4 \pm 1.88e\mbox{-}4$ & $3.23e\mbox{-}3 \pm 8.47e\mbox{-}4$ & $7.42e\mbox{-}3 \pm 7.40e\mbox{-}3$ \\
                         & 10.0   & \color{red}{$1.49e\mbox{-}4 \pm 8.74e\mbox{-}5$}   & $4.24e\mbox{-}3 \pm 3.18e\mbox{-}3$ & $1.73e\mbox{-}3 \pm 1.08e\mbox{-}3$ & $2.07e\mbox{-}3 \pm 8.84e\mbox{-}4$ \\
\\
\multirow{4}{*}{sin-high}&2.0     & $7.07e\mbox{-}1 \pm 6.70e\mbox{-}6$  & $5.91e\mbox{-}1 \pm 6.32e\mbox{-}3$ & $4.00e\mbox{-}2 \pm 1.24e\mbox{-}3$ & $2.23e\mbox{-}2 \pm 1.38e\mbox{-}3$  \\ 
                         & $\pi$  &  $7.07e\mbox{-}1 \pm 7.94e\mbox{-}6$ & $2.18e\mbox{-}1 \pm 1.88e\mbox{-}2$ & $3.06e\mbox{-}2 \pm 2.80e\mbox{-}3$ & $1.75e\mbox{-}2 \pm 2.44e\mbox{-}3$  \\ 
                         & 6.0    & $6.98e\mbox{-}1 \pm 4.88e\mbox{-}3$  & $1.44e\mbox{-}2 \pm 1.34e\mbox{-}3$ & $1.50e\mbox{-}2 \pm 8.68e\mbox{-}4$ & $7.03e\mbox{-}3 \pm 1.95e\mbox{-}3$  \\ 
                         & 10.0   & $6.58e\mbox{-}1 \pm 3.32e\mbox{-}3$  & $7.55e\mbox{-}3 \pm 1.73e\mbox{-}3$ & $1.12e\mbox{-}2 \pm 2.75e\mbox{-}3$ & \color{red}{$4.60e\mbox{-}3 \pm 3.70e\mbox{-}4$}  \\ 
\end{tabular}
\end{adjustbox}
\end{center}
\end{table}

\begin{table}[ht]
\caption{RMSE for different $\{h_i\}$ with exponential decay}
\label{table: sin-low with exponential decay}
\begin{center}
\begin{adjustbox}{width=\columnwidth, center}
\begin{tabular}{lllll}
\multicolumn{1}{c}{\bf Function name}& $h_0$ $\setminus$ M& \multicolumn{1}{c}{1} &\multicolumn{1}{c}{2} & \multicolumn{1}{c}{3}
\\ \hline \\
\multirow{4}{*}{sin-low} &2.0     & $7.80e\mbox{-}4 \pm 7.96e\mbox{-}4$   & $7.06e\mbox{-}4 \pm 2.54e\mbox{-}4$ & $6.33e\mbox{-}4 \pm 1.38e\mbox{-}4$  \\
                         & $\pi$  & $1.60e\mbox{-}4 \pm 6.66e\mbox{-}5$   & $5.40e\mbox{-}4 \pm 2.05e\mbox{-}4$ & $8.02e\mbox{-}4 \pm 6.58e\mbox{-}5$  \\
                         & 6.0    & $2.91e\mbox{-}4 \pm 1.22e\mbox{-}4$   & $1.03e\mbox{-}3 \pm 3.78e\mbox{-}4$ & $3.47e\mbox{-}3 \pm 3.31e\mbox{-}4$  \\
                         & 10.0   & \color{red}{$1.49e\mbox{-}4 \pm 8.74e\mbox{-}5$}   & $1.29e\mbox{-}3 \pm 1.74e\mbox{-}4$ & $3.30e\mbox{-}3 \pm 2.80e\mbox{-}4$  \\
\\
\multirow{4}{*}{sin-high} &2.0     & $7.07e\mbox{-}1 \pm 6.70e\mbox{-}6$    & $7.07e\mbox{-}1 \pm 1.42e\mbox{-}5$ & $7.02e\mbox{-}1 \pm 2.84e\mbox{-}3$ \\
                         & $\pi$  & $7.07e\mbox{-}1 \pm 7.94e\mbox{-}6$    & $6.89e\mbox{-}1 \pm 4.82e\mbox{-}3$ & $2.29e\mbox{-}2 \pm 1.76e\mbox{-}3$ \\
                         & 6.0    & $6.98e\mbox{-}1 \pm 4.88e\mbox{-}3$    & \color{red}{$1.20e\mbox{-}2 \pm 1.35e\mbox{-}3$} & $2.17e\mbox{-}2 \pm 6.67e\mbox{-}4$ \\
                         & 10.0   & $6.58e\mbox{-}1 \pm 3.32e\mbox{-}3$    & $7.22e\mbox{-}2 \pm 8.12e\mbox{-}3$ & $3.48e\mbox{-}2 \pm 2.38e\mbox{-}3$ \\
\end{tabular}
\end{adjustbox}
\end{center}
\end{table}
\section{Oscillations of SincKAN} \label{Appendix: oscillation}
\begin{figure}[ht]
  \centering 
  \subfigure[\label{2_8_1} $h_0=2.0, N=8, M=1$]{\includegraphics[width=0.4\linewidth]{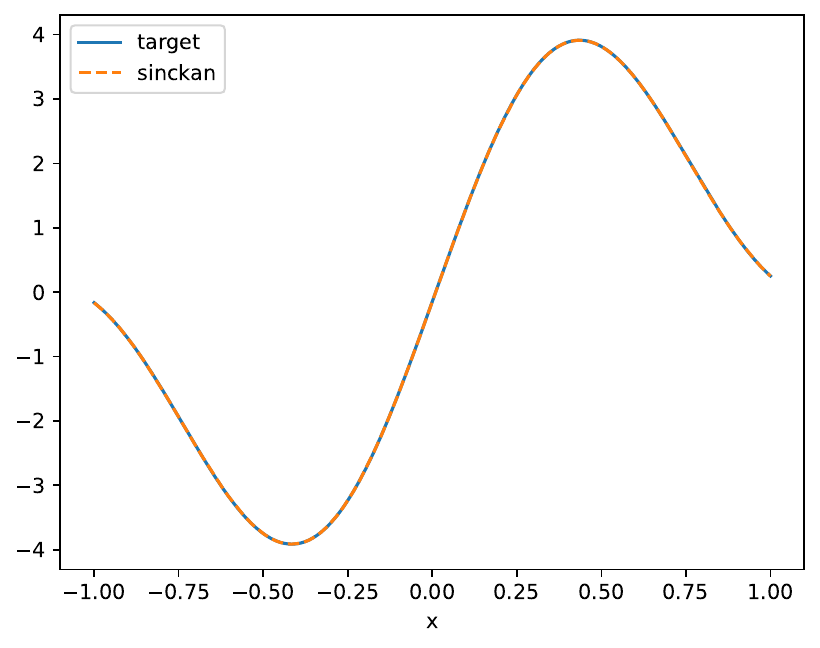}} 
  \subfigure[\label{2_8_6} $h_0=2.0, N=8, M=6$]{\includegraphics[width=0.4\linewidth]{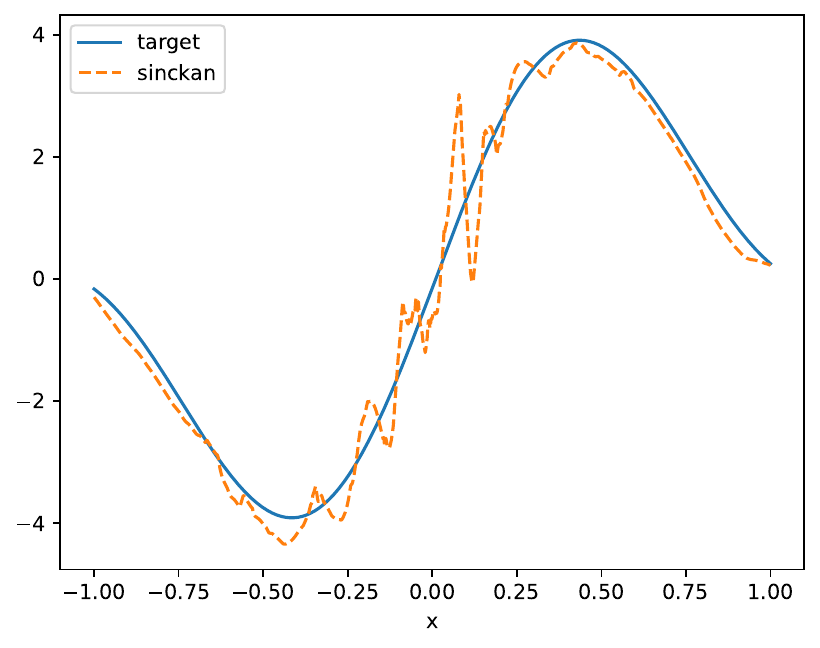}}
  \subfigure[\label{10_8_1} $h_0=10.0, N=8, M=1$]{\includegraphics[width=0.4\linewidth]{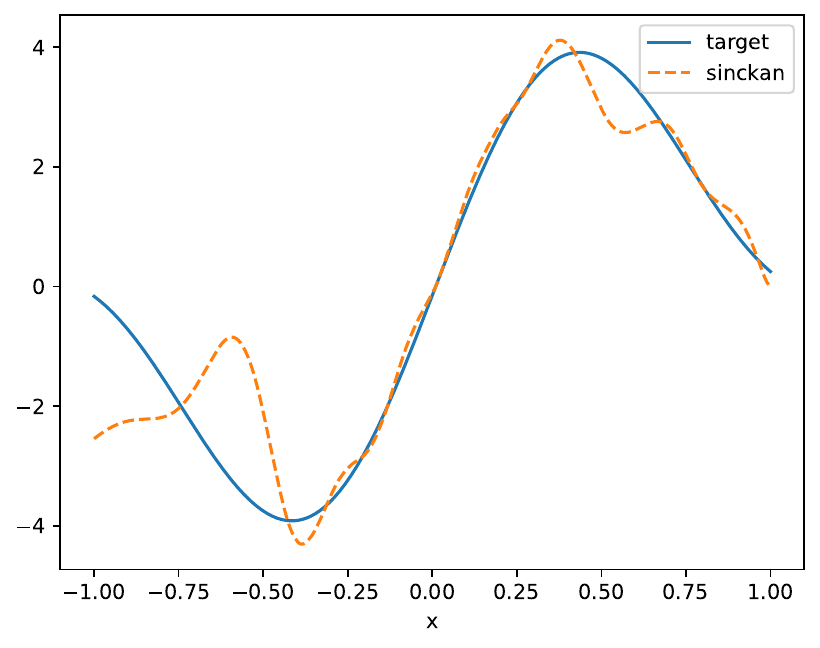}}
  \subfigure[\label{2_100_6} $h_0=2.0, N=100, M=6$]{\includegraphics[width=0.4\linewidth]{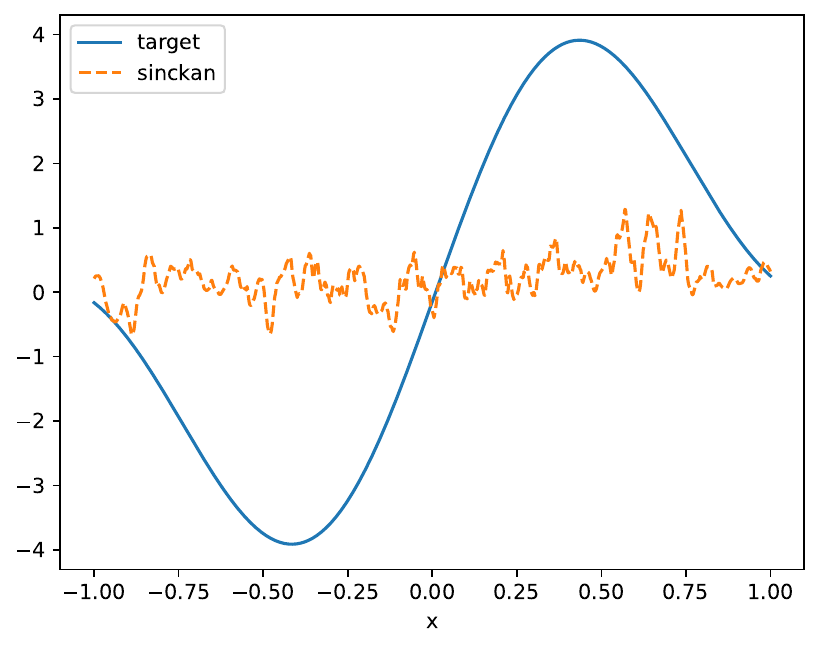}}
  \caption{\cref{2_8_1} solve \cref{eq:cd} accurately. However, SincKANs have oscillations after either increasing $M$ (\cref{2_8_6}) or increasing $h_0$. \cref{2_100_6} shows that with the same hyperparameters used in approximation, SincKAN becomes extremely inaccurate due to the violent oscillations.}
  \label{fig:oscillation}
\end{figure}

We conducted the experiments on convection-diffusion equations (\cref{eq:cd}) with $h_0=2.0,10.0$, $N=8,100$, and $M=1,6$. Except $h_0=2.0$ and $N=8$, the inaccuracy of derivatives makes SincKAN unstable with the loss diverging. We choose some figures plotted in \cref{fig:oscillation} to show the oscillations that limit the improvement of SincKANs.

\section{Metrics}
In this paper, we use two metrics. For interpolation, we inherit the RMSE metric from KAN~\cite{liu2024kan}, the formula is :
\begin{equation}
    \mathrm{RMSE}=\sqrt{\frac{1}{N}\sum_{i=1}^{N}(y_i-\hat{y}_i)^2};
\end{equation}
for PIKANs, we utilize the relative L2 error which is the most common metric used in PINNs:
\begin{equation}
    \mathrm{Relative L2}=\frac{\|\boldsymbol{y}-\hat{\boldsymbol{y}}\|_2}{\|\boldsymbol{y}\|_2},
\end{equation}
where $y$ is the target value, and $\hat{y}$ is the predicted value

\end{document}